\documentclass{article}

\usepackage{arxiv}

\usepackage[utf8]{inputenc} 
\usepackage[T1]{fontenc}    
\usepackage{hyperref}       
\usepackage{url}            
\usepackage{booktabs}       
\usepackage{amsfonts}       
\usepackage{nicefrac}       
\usepackage{microtype}      
\usepackage{lipsum}
\usepackage{fancyhdr}       
\usepackage{graphicx}       
\graphicspath{{media/}}     
\usepackage{booktabs}
\usepackage{multirow}
\usepackage[utf8]{inputenc}
\usepackage{amsfonts}
\usepackage{nicefrac}
\usepackage{microtype}
\usepackage{subcaption}
\usepackage{tikz}
\usepackage{centernot}
\DeclareMathAlphabet{\mathcal}{OMS}{cmsy}{m}{n}
\usepackage{setspace}
\usepackage{pdflscape}
\usepackage{colortbl}
\usepackage{textcomp}
\usepackage{float}
\usepackage{enumitem}
\usepackage{mathtools}
\usepackage{amsthm}
\usepackage{thmtools, thm-restate}
\usepackage{cleveref}
\usepackage{todonotes}
\usepackage{dashbox}
\usepackage{tabularray}
\usepackage{bbm}
\usepackage{etoolbox}
\usepackage{amsthm}
\usepackage{amsmath}
\newtheorem*{remark}{Remark}
\newtheorem*{proposition}{Proposition}
\newtheorem*{corollary}{Corollary}
\newtheorem*{definition}{Definition}
\newtheorem*{lemma}{Lemma}

\usepackage[round]{natbib}
\newcommand{\N}{\mathcal{N}}

\newcommand{\cf}{\scalebox{.4}{\textbf{CF}}}
\newcommand{\CF}{{\textbf{\small CF}}}
\newcommand{\Plus}{\raisebox{.4\height}{\scalebox{.6}{+}}}

\makeatletter
\newcommand*\bigcdot{\mathpalette\bigcdot@{.5}}
\newcommand*\bigcdot@[2]{\mathbin{\vcenter{\hbox{\scalebox{#2}{$\m@th#1\bullet$}}}}}
\makeatother
\newcommand\dboxed[1]{\dbox{\ensuremath{#1}}}

\title{Robustness Implies Fairness in Causal Algorithmic Recourse}

\author{
  Ahmad-Reza Ehyaei \\
  Department of Computer Science University of T\"ubingen \\
  T\"ubingen, Germany\\
  \texttt{ahmadreza.ehyaei@uni-tuebingen.de} \\
   \And
  Amir-Hossein Karimi \\
Max Planck Institute for Intelligent Systems \\
T\"ubingen, Germany\\
\texttt{amir@tue.mpg.de} \\
   \And
Bernhard Sch\"olkopf \\
Max Planck Institute for Intelligent Systems \\
T\"ubingen, Germany\\
\texttt{bernhard.schoelkopf@tuebingen.mpg.de} \\
   \And
Setareh Maghsudi \\
  Department of Computer Science University of T\"ubingen \\
T\"ubingen, Germany\\
\texttt{setareh.maghsudi@uni-tuebingen.de} 
}

\begin{document}
\maketitle

\begin{abstract}
Algorithmic recourse aims to disclose the inner workings of the black-box decision process in situations where decisions have significant consequences, by providing recommendations to empower beneficiaries to achieve a more favorable outcome. To ensure an effective remedy, suggested interventions must not only be low-cost but also robust and fair. This goal is accomplished by providing similar explanations to individuals who are alike.
This study explores the concept of individual fairness and adversarial robustness in causal algorithmic recourse and addresses the challenge of achieving both.
To resolve the challenges, we propose a new framework for defining adversarially robust recourse. The new setting views the protected feature as a pseudometric and demonstrates that individual fairness is a special case of adversarial robustness.
Finally, we introduce the fair robust recourse problem to achieve both desirable properties and show how it can be satisfied both theoretically and empirically.
\end{abstract}


\section{Introduction}
\label{sec:introduction}
Algorithmic Recourse refers to the capability of an algorithm to provide explanations and make recommendations in response to an appeal or challenge raised by an individual who has been affected negatively by its decision ~\cite{wachter2017counterfactual, ustun2019actionable, karimi2020algorithmic, venkatasubramanian2020philosophical}.
This concept is particularly important in areas such as finance, healthcare, and criminal justice where decisions made by algorithms can have significant impacts on people's lives~\cite{chou2022counterfactuals}. 
Recently, there has been an explosion of proposals for counterfactual explainers in the emerging field of algorithmic recourse
~\cite{guidotti2022counterfactual, stepin2021survey, karimi2021survey, verma2020counterfactual}. 

Ensuring fairness and robustness in algorithmic decision-making processes is crucial to guarantee fair and just outcomes for all involved.
In the context of algorithmic recourse, robustness refers to the ability of an algorithm to withstand unreliability, manipulation, or deception by malicious actors, while still providing fair and accurate recourse recommendations
\cite{slack2021counterfactual, upadhyay2021towards, dominguez2022adversarial, pawelczyk2022exploring}. 
There are four types of unreliabilities in counterfactual explanations \cite{mishra2021survey}:
\begin{itemize}
	\item Robustness to input perturbations: Examining recourse behavior in response to slight input changes while the classifier remains unchanged~\cite{dominguez2022adversarial}.
	\item Robustness to the uncertainty of SCM: There are no guarantees of recourse in structural causal models for unknown structural equations \cite{karimi2020algorithmic}.
	\item Robustness to model changes: Investigating recourse with unchanged inputs while altering the underlying classifier~\cite{ferrario2022robustness, pmlr-v124-pawelczyk20a}.
	\item Robustness to hyperparameter selection: Examining recourse sensitivity to variations in algorithmic recourse model's hyperparameters~\cite{dandl2020multi}.
\end{itemize}
In addition to robustness, fairness is a critical component of responsible algorithmic recourse design and deployment. Fairness aims to ensure that the algorithm does not discriminate against certain groups of people based on sensitive characteristics such as race, gender, or age~\cite{mehrabi2021survey}. 
To prevent individuals from experiencing disparate treatment from algorithmic decisions, various criteria have been proposed ~\cite{mehrabi2021survey} for assessing the fairness of not only predictions but also recourse offered either with regards to independently changeable features~\cite{gupta2019equalizing} or those under assumptions of causal relationships~\cite{von2022fairness}.
\paragraph{Related Works}
There has been little research in the field of causal algorithmic recourse that addresses either fairness or robustness considerations.  
In the study conducted by \citet{von2022fairness}, a causal framework for defining fairness in algorithmic recourse was proposed, and it was argued that individual-level unfairness in a causal structure is better than group-level measurements.
On the other hand, \citet{dominguez2022adversarial} proposed a setting for defining perturbation and robustness under causal conditions and defined an adversarially robust recourse problem where the recourse is robust under small input perturbations.

\citet{artelt2022explain} presents a group fair counterfactual explanation algorithm under the non-causal framework.
In \citet{garg2019counterfactual}, a method for ensuring fairness in text classification tasks is proposed. The authors suggest using a robustness-based approach to achieve counterfactual fairness.
They employ a simple causal model for text generation and consider the counterfactual token fairness concept which is distinct from individual fairness \cite{counterfactual_fairness}.
By \citet{artelt2021evaluating}, a method for evaluating the robustness of counterfactual explanations (CFs) is presented. The authors propose that plausible CFs can be used instead of closest CFs to improve the robustness and individual fairness of algorithmic recourse.

By \citet{xu2021robust}, the authors investigate the relationship between robustness and fairness in the context of adversarial training. They find that while adversarial training can enhance the robustness of models, it can also result in a decrease in fairness. To address this trade-off, the authors propose the Fair-Robust-Learning (FRL) framework, which balances robustness and fairness by incorporating fairness constraints into the adversarial training process.
In the study by \citet{ustun2019actionable}, the authors evaluate recourse cost for the linear classification model, and its explicit formula help to analyze the various aspects of the recourse problem.
To the best of our knowledge, in the causal framework there have been no works that address the simultaneous need for both fair and robust recourse, and offer insight into the relations therein.
%
\paragraph{Contributions}
The current formulation of robust recourse lacks fairness, impacting the efficacy of algorithms in practical applications. Our study seeks to improve algorithmic recourse by incorporating both individual fairness and adversarial robustness.
To accomplish this, we suggest utilizing protected features as a pseudometric and incorporating counterfactual twins into the perturbation ball in order to embed fairness within robustness.
We highlight the following contributions:
\begin{itemize}
	\item Generalize the explicit recourse cost proposed by \citet{ustun2019actionable} to arbitrary $L_p$ norm and obtain an explicit formula for intervention, counterfactual twins, and adversarially robust recourse cost for linear SCM and classifier.
	\item Show fairness' impossibility theorem for the case of a linear classifier.
	\item A new definition of a protected group is proposed, utilizing the concept of a pseudometric.
	\item We generalize the approach proposed by \citet{dominguez2022adversarial} to \emph{adversarially fair robust} recourse problem.
	\item We demonstrate that individual fairness can be viewed as a specific case of robustness in the new recourse problem setting.
	\item We propose a new recourse problem incorporating both fairness and robustness.
\end{itemize}

\section{Preliminaries \& Background}
\label{sec:background}

\paragraph{Notation.}
Let $\textbf{V}$ be the observed random vector 
which contains $m$ \textbf{categorical} and $n$ \textbf{continuous} features $\textbf{V} = (\textbf{Z}, \textbf{X})$.
Continuous variables $\textbf{X}=(\textbf{X}_1, ..., \textbf{X}_n)$ take values over a real number $x=(x_1, \dots, x_n)\in \mathcal{X} = \mathcal{X}_1 \times ...  \times \mathcal{X}_n \subseteq \mathbb{R}^n$, 
and Similarly, \textbf{categorical} variables $\textbf{Z}=(\textbf{Z}_1, ..., \textbf{Z}_m)$ take values from a subset of an integer numbers $z=(z_1, ..., z_n) \in \mathcal{Z} = \mathcal{Z}_1\times ...  \times \mathcal{Z}_m \subseteq \mathbb{Z}^m$ such that $|\mathcal{Z}_i|< \infty$. 
Let $h:\mathcal{V} \rightarrow \mathcal{Y}$ be a given \textbf{binary classifier} with $y = h(v) \in \mathcal{Y} =\{\pm 1\}$ where a positive value indicates a favorable condition (for example, approve credit). The empirical dataset $\mathcal{D} = \{(\mathbf{v}_i,y_i) \}_{i=1}^N$ are i.i.d. samples of the random variables $\textbf{V}$ and $\textbf{Y}$.
\paragraph{Structural Causal Model.}
\label{sec:causality}
Suppose the observed r.v. $\mathbf{V}=\{\mathbf{V}_1, \ldots, \mathbf{V}_{n+m}\}$ is generated by the \textbf{structural causal model} (SCM) $\mathcal{M}=(\mathbb{S}_{\mathcal{M}}, \mathbb{P}_{\mathbf{U}})$~\cite{pearl2009causality},
with the \textbf{structural equations}
$\mathbb{S}_{\mathcal{M}} = \left\{\mathbf{V}_i := f_i\left(\mathbf{V}_{\text{pa}(i)}, \mathbf{U}_i\right)\right\}_{i=1}^{n+m}$
which describes the causal relationship between any endogenous variable $\mathbf{V}_i$, its direct causes $\mathbf{V}_{\text{pa}(i)}$ and an exogenous variable $\mathbf{U}_i$ by means of the deterministic function $f_i$. 
We assume that $\mathcal{M}$ is {causally sufficient}, which means that the distribution $\mathbb{P}_{\mathbf{U}}$ factorizes on the latent variables $\mathbf{U}=\{U_i\}_{i=1}^{n+m}$.
When the causal graph is acyclic, the distribution $\mathbb{P}_{\mathbf{U}}$ implies a unique \textbf{push-forward distribution} $\mathbb{P}_{\mathbf{V}}$ over the features $\mathbf{V}$. 
The structural equations $\mathbb{S}_{\mathcal{M}}$ also induce a mapping $\mathbf{S}_{\mathcal{M}}: \mathcal{U} \rightarrow \mathcal{V}$ between exogenous and endogenous variables and inverse image
$\mathbf{S}_{\mathcal{M}}^{-1}: \mathcal{V} \rightarrow \mathcal{U}$ such that $\mathbf{S}_{\mathcal{M}}\left(\mathbf{S}_{\mathcal{M}}^{-1}(v)\right)=v$ for all $v\in\mathcal{V}$.
An \textbf{Additive noise model (ANM)}~\cite{hoyer2008nonlinear} is a class of invertible SCMs, where the structural equations of $\mathbb{S}_{\mathcal{M}}$ and its inverse have a form:
\begin{equation}
\label{eq:ANM}
\mathbb{S}_{\mathcal{M}} = \{\textbf{V}_i := f_i(\textbf{V}_{\mathbf{pa}(i)})+\textbf{U}_i \}_{i=1}^{n+m} \quad 
\implies  \quad  u_i=v_i-f_i(v_{\mathbf{pa}(i)}), \quad i \in \{1, 2, \dots, n+m\},
\end{equation}
An example of ANM is \textbf{linear} SCMs, which $f_i$ are considered linear functions.
\paragraph{Causal Intervention.}
SCMs can be used to study the impact of interventions, including external system manipulations that change the data generation process, and there are two types of interventions \cite{peters2017elements}. 
With \textbf{Hard intervention}s (with \textbf{do}-operator notation $\mathcal{M}^{do(\mathbf{V}_{\mathcal{I}}=\mathbf{\theta})}$), the feature values $\mathbf{V}_{\mathcal{I}}$ of a subset $\mathcal{I}\subseteq \{1, 2, \dots, n+m\}$ are fixed to some constant $\theta\in\mathbb{R}^{|\mathcal{I}|}$ by removing some parts of the structural equations: 
\begin{equation}
\label{eq:hard}
\mathbb{S}_{\mathcal{M}}^{do(\mathbf{V}_{\mathcal{I}}=\mathbf{\theta})}= 
\begin{cases}
\textbf{V}_i := \mathbf{\theta}_i & \text{if} \ i \in \mathcal{I} \\
\textbf{V}_i := f_i (\mathbf{V}_{\text{pa}(i)}, \mathbf{U}_i )& \text{otherwise}
\end{cases}
\end{equation}
Hard interventions break the causal relationship between the affected variables and all of their ancestors in the causal graph.
\textbf{Soft intervention}, on the other hand, preserves all causal relationships and changes only structural equation functions. 
For example, \textbf{additive interventions} \cite{eberhardt2007interventions} with symbol $\mathcal{M}^{do(\mathbf{V}_{\mathcal{I}} = v_{\mathcal{I}} + \mathbf{\delta})}$, 
\footnote{In the causality literature, the do-operator is only applied to hard interventions. In this work, to avoid using more notation, we use the $do(\mathbf{V}_{\mathcal{I}} = v_{\mathcal{I}} + \mathbf{\delta})$ for additive interventions too.}
were changed the features $\mathbf{V}$ by some perturbation vector $\delta\in\mathbb{R}^{n+m}$: 
\begin{equation}
\label{eq:soft}
\mathbb{S}_{\mathcal{M}}^{do(\mathbf{V}_{\mathcal{I}} = v_{\mathcal{I}} + \mathbf{\delta})} = \left\{V_i := f_i\left(\mathbf{V}_{\text{pa}(i)}, \mathbf{U}_i\right)+\delta_i\right\}_{i=1}^{n+m}.
\end{equation}
is an example of soft intervention.
For additional information and examples of additive interventions, refer to section 4.4~\citet{glymour2016causal}.
\paragraph{Counterfactuals.}
SCMs enable the examination of counterfactual statements and the reasoning of outcomes under hypothetical interventions on a variable.
In order to determine the counterfactual of instance $v$, represented by $v^{\text{CF}}$, the following steps should be taken: (1) identify the exogenous variables $u$ that correspond to $v$ and (2) employ the modified structural equations to $u$.
For hard and additive interventions, respectively we denote the corresponding counterfactual map
$v_{\theta}^{\text{CF}} := \mathbf{CF}(v, do(\mathbf{V}_{\mathcal{I}} = \mathbf{\theta});\mathcal{M})=\mathbf{S}_{\mathcal{M}}^{\mathbf{\theta}}(\mathbf{S}_{\mathcal{M}}^{-1}(v))$
and
$v_{\delta}^{\text{CF}} := \mathbf{CF}(v, do(\mathbf{V}_{\mathcal{I}} = v_{\mathcal{I}} + \mathbf{\delta}); \mathcal{M})=\mathbf{S}_{\mathcal{M}}^{\delta}(\mathbf{S}_{\mathcal{M}}^{-1}(v))$
where $\mathbf{S}_{\mathcal{M}}^{\theta}$  and $\mathbf{S}_{\mathcal{M}}^{\delta}$ are simpler notation instead of 
$\mathbf{S}_{\mathcal{M}}^{do(\mathbf{V}_{\mathcal{I}}=\mathbf{\theta})}$ and
$\mathbf{S}_{\mathcal{M}}^{do(\mathbf{V}_{\mathcal{I}}= v_{\mathcal{I}} + \mathbf{\delta})}$ respectively. 
From now we use the notations 
$\mathbf{CF}(v, \mathbf{\theta};\mathcal{M})$
and
$\mathbf{CF}(v, \mathbf{\delta};\mathcal{M})$
instead of
$\mathbf{CF}(v, do(\mathbf{V}_{\mathcal{I}} = \mathbf{\theta});\mathcal{M})$ 
and
$\mathbf{CF}(v, do(\mathbf{V}_{\mathcal{I}} = v_{\mathcal{I}} + \mathbf{\delta}); \mathcal{M})$
for simplicity.
\paragraph{Recourse Problem.}
The causal recourse problem \cite{karimi2020algorithmic} involves finding the minimum cost feasible intervention by taking into account actionability constraint that would positively classify the corresponding counterfactual. For the hard intervention, the corresponding optimization problem is:
\begin{equation*} 
\label{eq:hard_rec}
a^*_{\theta}(v) = \underset{a = do(\mathbf{V}_\mathcal{I} = \theta) \in \mathcal{F}(v)}{\textrm{argmin}}
cost(v, a)
\quad \text{s.t.}
\quad
h(v_{\theta}^{\text{CF}})=1 
\end{equation*}
where $\mathcal{F}(v)$ is the set of feasible actions from factual instance $v$.
As a hard intervention on all features would eliminate the relationship between counterfactual $\mathbb{CF}(v, a; \mathcal{M})$ and $v$, the \citet{dominguez2022adversarial} rewrote the recourse problem using additive intervention to avoid this issue. 
Assuming $a^*_{\theta}(v)$ and $a^*_{\delta}(v)$ are a minimal cost action of recourse problem w.r.t. hard and additive intervention, then the \textbf{recourse cost} are defined as
$r_{\theta}^{\mathcal{M}}(v) = cost(v, a^*_{\theta}(v))$
and
$r_{\delta}^{\mathcal{M}}(v) = cost(v, a^*_{\delta}(v))$.
\paragraph{Dist and Cost.}
\footnote{Metric is commonly called the dissimilarity function and is written as \textbf{dist}.}
The \textbf{cost} function, which measures the expense of altering an individual's attributes, is a crucial aspect of algorithmic recourse. Each person may have a distinct cost function \cite{venkatasubramanian2020philosophical}.
However, the \textbf{dissimilarity} function (dist) is distinct from the concept of cost, it aims to evaluate the level of difference between instances from a specific perspective.
For example, in a drug study, the cost of changing one person's eye color to different shades would be infinite (impossible), while the dist between two people's eye color may be minimal from a clinical perspective. There is usually no direct correlation between cost and dissimilarity functions~\cite{karimi2021survey}.
\paragraph{Adversarial Robust Recourse Problem.}
To Add robustness property to causal recourse, \citet{dominguez2022adversarial} proposed a method that recourse recommendations should remain valid under small perturbation of input instance.
For instance $v$, SCM $\mathcal{M}$ and norm $\|\cdot\|$, the \textbf{additive counterfactual perturbation (ACP)} $B^{\mathcal{M}}_{\Delta \Plus}(v)$ with radius $\Delta \geq 0$  is defined as the set of causal counterfactuals under $\delta$-additive interventions:
\begin{equation}
\label{eq:ACP}
B^{\mathcal{M}}_{\Delta \Plus}(v) = \{\dboxed{\mathbf{CF}(v, \delta; \mathcal{M})} \; : \; \dboxed{\|\delta\| \leq \Delta} \}
\end{equation}
\begin{figure}[ht]
\centering
\begin{subfigure}{0.32\textwidth}
\includegraphics[width=\textwidth]{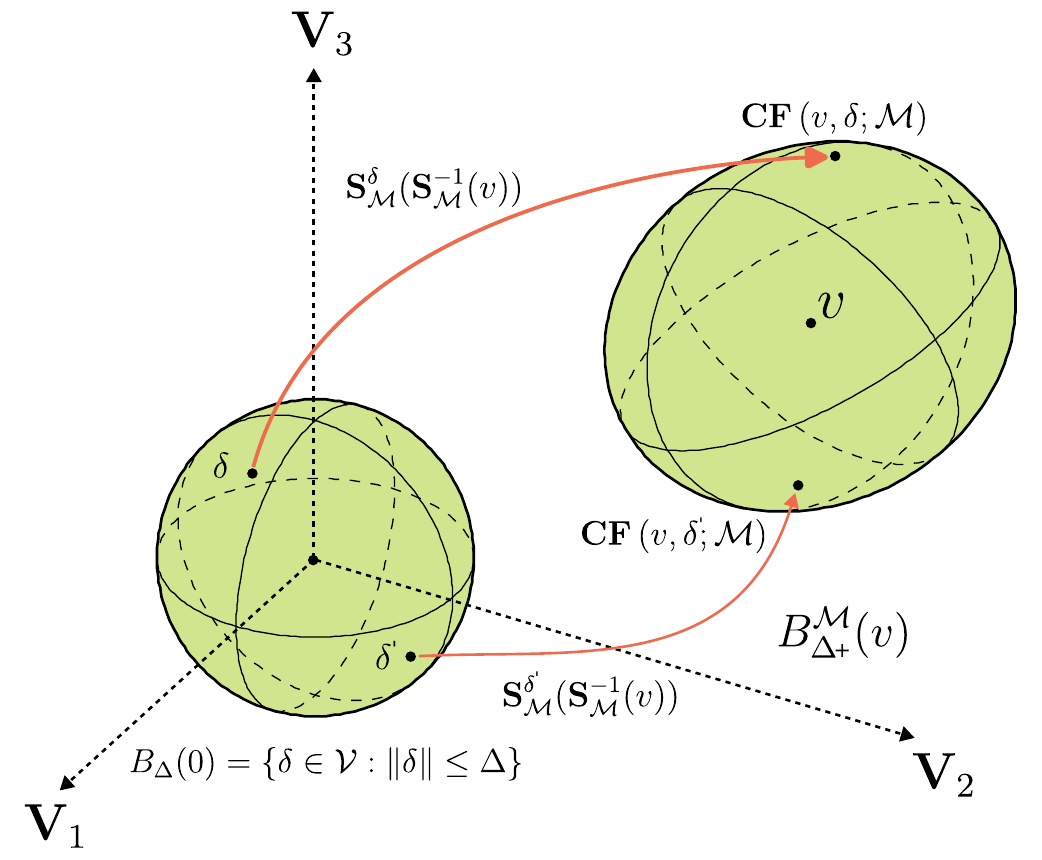}
 \caption{}
 \end{subfigure}
\begin{subfigure}{0.32\textwidth}
\includegraphics[width=\textwidth]{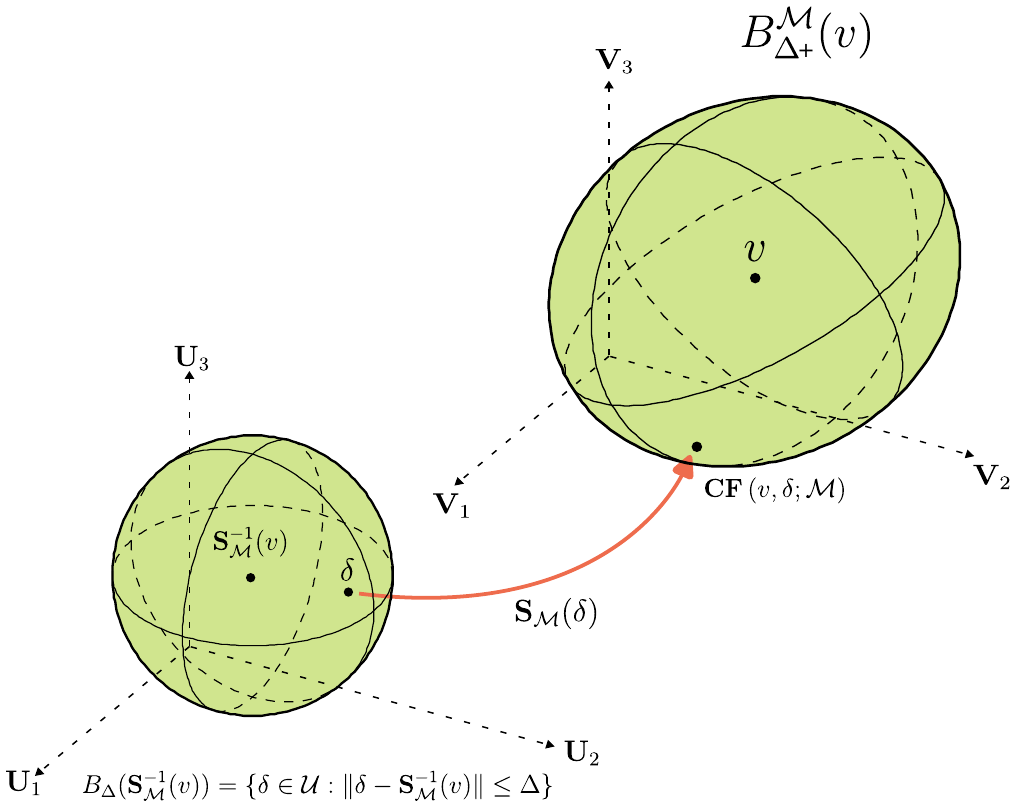}
\caption{}
\end{subfigure}
\begin{subfigure}{0.32\textwidth}
\includegraphics[width=\textwidth]{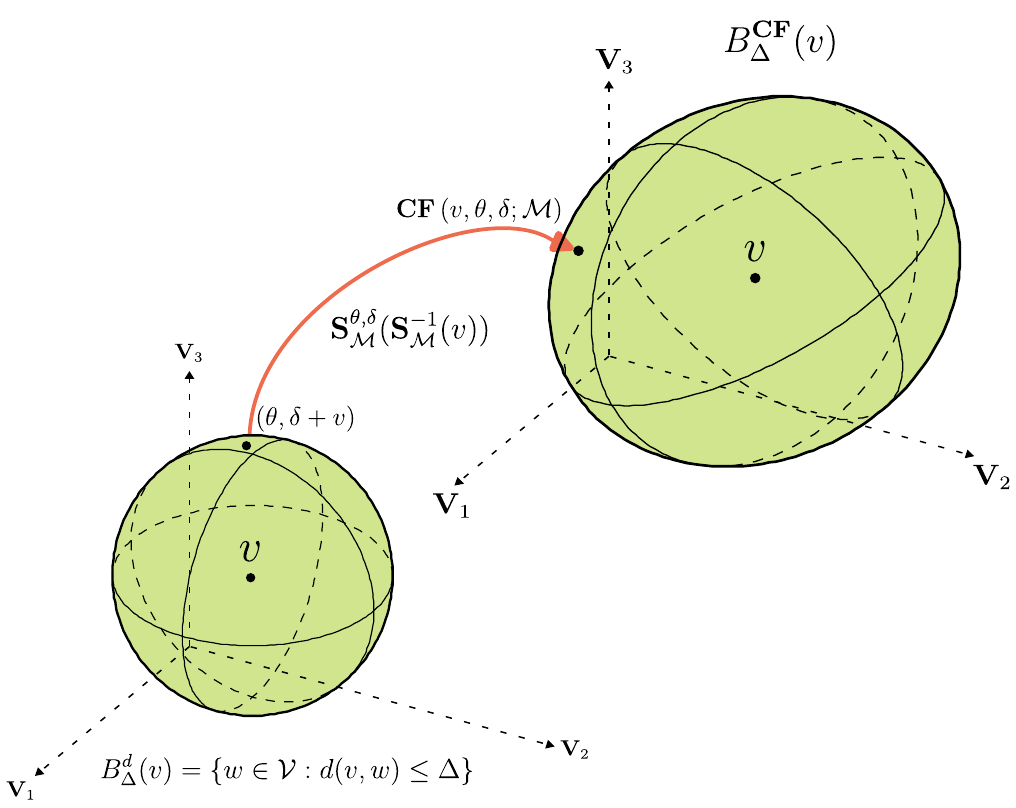}
\caption{}
\end{subfigure}
\caption{\textbf{Intuition of Counterfactual Ball:} 
 (a) The geometric interpretation of the ACP is that for each point of $\delta \in B_{\Delta}(0)$ constructs the $\mathbf{S}_{\mathcal{M}}^{\delta}$ and compute counterfactual explanation of $v$ respect to it.
 (b) In ANM SCMs the ACP can be considered as a map between unit sphere in $\mathcal{U}$ to $\mathcal{V}$ by $\mathbf{S}_{\mathcal{M}}$.
 (c) The definition of the counterfactual ball with metric and additive intervention
 }
	\label{fig:acp}
\end{figure}
Additive counterfactual perturbation has a complicated geometric interpretation. It needs to know the map $\mathbf{S}_{\mathcal{M}}^{\delta}$ for each perturbation $\delta$ (see Fig. \ref{fig:acp} (a). 
But in ANMs, ACP can be considered as a mapping of perturbation ball in exogenous space to ACP in endogenous space by means of the map $\mathbf{S}_{\mathcal{M}}$ (see Fig. \ref{fig:acp} (b)). 
This means that if we set $u = \mathbf{S}_{\mathcal{M}}^{-1}(v)$, we can define a perturbation ball on the exogenous space as $B_{\Delta}(u) = \{ u_i + \delta_i \ : \ \|\delta\| \leq \Delta\}$ around $u$, then ACP equals to:
\begin{equation}
\label{eq:ACPB}
B^{\mathcal{M}}_{\Delta \Plus}(v) = \mathbf{S}_{\mathcal{M}}(B_{\Delta}(u))   
\end{equation}
The \textbf{robust recourse problem} for a specific additive counterfactual perturbation
$B^\mathcal{M}_{\Delta \Plus}(v)$, involves finding the most cost-effective recourse action that is resistant to adversarial perturbation, i.e.
\begin{equation}
\label{def:robrec}
 	a_{\Delta \Plus}^*(v) =  \underset{a \in \mathcal{F}(v)}{\textrm{argmin}}{} \;  cost(v, a) \quad
 	\text{s.t} \quad  h(\mathbb{CF}(v', a;  \mathcal{M}))=1 \quad \forall v' \in B^\mathcal{M}_{\Delta \Plus}(v)  
\end{equation}
\paragraph{Fair Recourse Problem.}
A \textbf{protected group} (or variable) is a group that should not be subject to discrimination~\cite{verma2018fairness}, and \textbf{fairness} is the concept that users within a protected group should be treated similarly by models.
Let $\mathbf{A} \in \mathbf{Z}$ is protected variable that has finite levels $\mathcal{A} = \{a_1, \dots, a_k\}$.
For each instance $v$ of $\mathcal{M}$, the set of \textbf{counterfactual twins} w.r.t protected variable $A$ is defined as
$\ddot{\mathbbm{v}} = \{ \ddot{v}_a = \mathbf{CF}(v,do(A=a); \mathcal{M}) :   a \in \mathcal{A} \}$. 
For convenience, the twin mapping $\mathbf{S}_{\mathcal{M}}^{do(A=a)}$ according $\mathcal{M}^{do(A=a)}$ is denoted by $\ddot{\mathbf{S}}_a$.
The \textbf{individual-level unfairness} of causal recourse for a dataset $\mathcal{D}$, classifier $h$, and cost function $c$ \textit{w.r.t.\ an SCM} $\mathcal{M}$ is defined as:
\begin{equation}
\label{eq:indfair}
\sigma_\mathrm{ind}(\mathcal{D}, h, c, \mathcal{M})
:=
\max_{a\in\mathcal{A}; v\in\mathcal{D}}
\left|
r^{\mathcal{M}}(v) - r^{\mathcal{M}}(\ddot{v}_a))
\right|
\end{equation}
Recourse is \textbf{individually fair} if~$\sigma_\mathrm{ind}=0$ \cite{von2022fairness, gupta2019equalizing}.
The recourse fairness is highly influenced by the fairness of the classifier.
A classifier $h$ w.r.t protected variables $A$ an instance $v$ of SCM $\mathcal{M}$ is \textbf{counterfactually fair} if satisfies $h(v) = h(\ddot{v}_a)$ for all $a \in \mathcal{A}$.
%
%
\section{The Fair Robust Recourse Challenges}
\label{sec:challenges}
In this section, we aim to identify obstacles to the attainment of fair and robust recourse. To effectively illustrate the bottlenecks that require alteration, we endeavor to derive explicit formulas for counterfactuals, twins, recourse costs, and adversarial recourse costs. To achieve this, we limit our analysis to linear SCMs with linear classifiers. 
Additionally, we assume that $\mathcal{M}$, has one protected categorical feature and the remaining variables are continuously manipulable. Furthermore, we consider the distance function to be $L_q$ where $0 < q \leq \infty$.
In line with the work of \citet{wachter2017counterfactual,ustun2019actionable}, we employ a cost function that depends on instance $v$ and structure of $\mathcal{M}$, in order to identify optimal counterfactual explanations. 
Therefore for action $a = do(X_{\mathcal{I}}= x_{\mathcal{I}} + \delta)$, we consider $cost(v,a) = \|v-\CF(v,a;\mathcal{M}) \|_p$
rather than $cost(v,a) = \|\delta \|$ as proposed in the work of ~\citet{karimi2020algorithmic}. The complete set of assumptions is outlined in Tab. \ref{table:condition}.

While the conditions outlined in this section may be expanded upon in a more general manner, the complexity of the resulting formulas precludes us from achieving our objective of identifying fair and robust recourse. The generalization of the concepts presented in this chapter will be addressed in future research. 
We demonstrate that despite the underlying assumptions are not complex, the implementation of fairness in SCMs faces a significant challenge.
\SetTblrInner{rowsep=4pt,colsep=3pt}
\begin{table}[ht]
\centering
\begin{tblr}{ | c  p{14cm} |}
\hline
\textbf{Notion} & \textbf{Condition} \\
\hline
$\mathcal{M}$ & Linear SCM with variables $V = (A,X_1, \dots, X_n)$. \\ 
$A$ & Protected variable with levels $\mathcal{A} = \{a_1,\dots,a_k \}$ where $\{a_i\}_{i = 1}^{k}$ are increasing integer sequence. \\
$X_i$ & Real-valued continuous manipulable variables. \\ 
cost & Cost function $cost(v,a) = \|v-v^{\cf}_a \|_p$  where $0 < p \leq \infty$ where $v^{\cf}_a = \CF(v,a\mathcal{M})$. \\
dist & Dissimilarity function $dist(v,v') = \|v-v' \|_{q}$  where $0 < q \leq \infty$. \\
$h$ &  Linear classifier $h(v) = \text{sign}(w^T \bigcdot v - b)$ where $w = (w_0, w_1,..., w_n) \in \mathbb{R}^{n+1}$. \\ \hline
\end{tblr}
\caption{The assumptions utilized for deriving explicit formulas for the recourse problem.}
\label{table:condition}
\end{table}

The cornerstone of causal algorithmic recourse is the counterfactual computation of instance  $v$. In the following proposition, the hard and additive counterfactuals formulas can be found.
\begin{proposition} \label{pr:cfc}
Let $\mathcal{M}$ be SCM with conditions of Tab. \ref{table:condition} and inter-space map $S$, then the counterfactuals at instance $v$ have the following formula:
\begin{enumerate}[label=(\alph*)]
	\item The additive intervention $do(V = v + \delta)$  is:
	\begin{equation}
		v_{\delta}^{\text{CF}} = \CF(v,\delta;\mathcal{M}) = v + S\times \delta
	\end{equation}
	\item The hard intervention $do(V_i = \theta)$  is:
	\begin{equation} \label{eq:hardInterv}
	v_{\theta}^{\text{CF}} = \CF(v,\theta;\mathcal{M}) = v + (\theta -S^{-1}(v)_i)  S_{*,i}
	\end{equation}
\end{enumerate}
where $\times$ is matrix product, $S_{*,i}$ is the $i$-th column of matrix $S$ and $S^{-1}(v)_i$ is $i$-th element of vector $S^{-1}(v)$.
\end{proposition}
\begin{figure}[ht]
\centering
\includegraphics[width=1\textwidth]{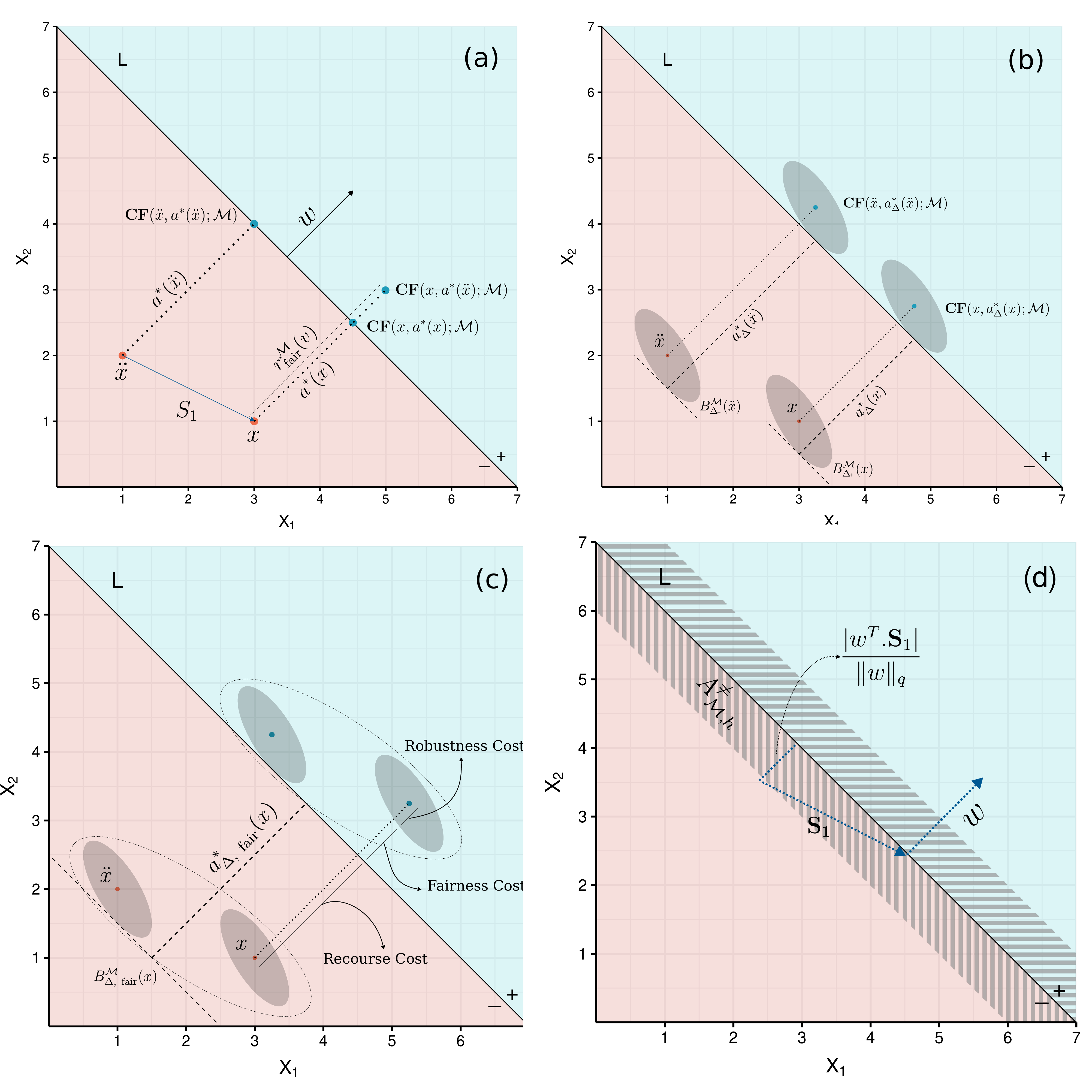}
\caption{\textbf{The Fair Robust Recourse Challenges}: (a) The minimum-cost action is equal to the minimum distance from continuous part $v$ to boundary line $L$. (b) Robust recourse cost for instance $v$ and its twin. (c) Additional costs for fairness and robustness to have fair robust recourse. (d) Unfair area for linear SCM and linear classifier.}
\label{fig:chellenge}
\end{figure}
For the $L_p$ cost function, the explicit formula for the recourse cost can be obtained following the intervention formula.
\begin{proposition} \label{pr:recdis}
Consider recourse problem with conditions of Tab. \ref{table:condition}, then the recourse cost w.r.t the linear classifier is given by the shortest distance (w.r.t. cost) of instance $v = (a,x)$ to classifier boundary line $L$:
\begin{enumerate}[label=(\alph*)]
\item If $A$ is immutable then:
\begin{equation}
\label{eq:recdis}
r^{\mathcal{M}}(v) = \dfrac{|w^T \bigcdot v-b|}{\|w\|_{p^*}} 
\end{equation}
where $p^*$ is conjugate of $p$ that satisfies equation $\frac{1}{p}+\frac{1}{p^*} = 1$ ($\|.\|_{p^*}$ is dual norm of $\|.\|_{p}$). 
\item If $A$ is mutable then:
\begin{equation}
\label{eq:mrecdis}
r^{\mathcal{M}}(v) = 
\min_{a' \in \mathcal{A}} \bigg\{\big(|a-a'|^p +  \dfrac{|w_0 a' +\sum_{i = 1}^{n} w_i x_i -b|^p}{\|w\|_{p^*}^p}\big)^{\frac{1}{p}} \bigg\}.
\end{equation}
\item
If $A$ is immutable, the optimal hard intervention is equal to $a_{\theta}^*(v) = do(V = v + (0, \eta))$ and respectively the optimal additive intervention is obtained by the formula $a_{\delta}^*(v) = do(V = v +   (0, \eta) \times (S^{-1})^T)$, where $v + (0, \eta)$ is a point on classifier boundary $L$ which has the minimum cost to instance $v$. Each element of $\eta_i$ is given by:
\begin{equation*}
\label{eq:cfair}
\eta_i = 
\begin{cases}
	- \dfrac{|w^T \bigcdot v-b|. |w_i|^{\frac{1}{p-1}}.sign(\frac{w^T \bigcdot v-b}{w_i})}{\|w\|_{p^*}}  & w_i \neq 0 \\
	0  & w_i = 0 \\
	\end{cases}
\end{equation*}
In case $A$ is mutable, let $a'$ be the level at which Eq. \ref{eq:mrecdis} is minimal. By substituting $b$ by $b - w_0.(a'-a) $ in the above formula the  $\theta$ and $\delta$ are achieved in similar immutable cases. 
\item If the cost function is weighted combinations of $L_p$ norms $\|.\| := \sum_{i = 1}^{n} \alpha_i\| .\|_{p_i}$, where $\sum_i \alpha_i = 1: \alpha_i \in \mathbb{R}^{\Plus}$  and $0 < p_i \leq \infty$, then recourse problem has the solution with cost that satisfies in the below inequality:
\begin{equation*}
\sum_{i = 1}^{n} \alpha_i r_{i}^{\mathcal{M}}(v) \leq r^{\mathcal{M}}(v) \leq r_{i_{*}}^{\mathcal{M}}(v)
\end{equation*}
where $r_{i}^{\mathcal{M}}(v)$ is the recourse cost corresponding to the norm $\|.\|_{p_i}$ and $r_{i_{*}}^{\mathcal{M}}(v)$ is the recourse cost of minimum norm with index $i_{*} =  \textrm{argmin}_{i}^{} \{ p_i\}$.
\end{enumerate}
\end{proposition}
Prop. \ref{pr:cfc} can be used to find the counterfactual twins of $v$, and the following corollary is the form of counterfactual w.r.t hard interventions of the different levels of protected variable $A$.
\begin{corollary}
\label{pr: twins}
Suppose $\mathcal{M}$ is SCM with conditions of Tab. \ref{table:condition}, so the counterfactual twins of $v$ is given by linear shift:
\begin{equation}
\label{eq: twins}
\ddot{\mathbbm{v}} = \{ \ddot{v}_{a'} = v+ (a' - a)      S_{*,1} : \   a' \in \{a_1, \dots, a_k \} \}
\end{equation}
\end{corollary}
The necessary condition of individually fair recourse is that the classifier should be counterfactually fair.
So, we need to make sure that the equation $h(v) = h(\ddot{v}_a)$ is true for all $a \in \mathcal{A}$. The instances that do not have this property are called belongs to \textbf{unfair area}.
\begin{definition}[Unfair Area]
Let $\mathcal{M}$ be an SCM and $h$ is a classifier for $\textbf{V}$, then the unfair area contains the instances where we have no counterfactual fair property:
\begin{equation}
	A_{\mathcal{M},h}^{\neq} := \{v \in \mathcal{V}: \exists a \in \mathcal{A} \quad \text{s.t.} \quad h(v) \neq h(\ddot{v}_a)\}
\end{equation}
\end{definition}
As seen in Fig. \ref{fig:chellenge} (d), the unfair area is a symmetric strip around  the decision boundary $L$.
The area $A_{\mathcal{M},h}^{\neq}$ can be determined explicitly for linear SCMs and linear classifiers.
\begin{proposition}
\label{pr:unfairEq}
By the conditions of Tab. \ref{table:condition},
the unfair area $A_{\mathcal{M},h}^{\neq}$, equals to the band that parallels to classifier boundary $L$:
\begin{equation}
\label{eq:unfairEq}
\{v \in \mathcal{V}: dist(v,L) \leq \underset{a,a' \in \mathcal{A}}{\max} \biggl\{ \dfrac{|(a' - a) \times w^T\bigcdot S_{*,1}|}{\|w\|_{p^*}} \biggl\}
\end{equation}
\end{proposition}
Linear SCMs with linear classifiers have individual fairness if they meet the following conditions.
\begin{proposition}
	\label{pr:linfair}
 
	Suppose we have a recourse problem with conditions of Tab. \ref{table:condition} and $A$ is immutable.
	Then there is no individually fair recourse for $\mathcal{M}$, unless $\mathbf{S}_{\mathcal{M}}$ and $w$ satisfy the equation
	$w^T \bigcdot S_{*,1} = 0$, that means $h$ only depends on a subset of variables that are non-descendants of $A$ in $\mathcal{M}$.
\end{proposition}
If $A$ is mutable, Eq. \ref{pr:linfair} does not have a simple form, so we omit this case since it is not relevant to subsequent topics. 
As shown in Prop. \ref{pr:linfair}, despite the linear conditions we do not have individual fairness w.r.t $A$ unless $w^T \bigcdot S_{*,1} = 0$.
This highlights the first challenge in defining the fair recourse problem.
\paragraph{Challenge I} 
\label{tx:cl1}
Unfairness affects both the decision-maker and the decision-subject. Regularization is a solution for reducing unfairness in decision-making models but can decrease prediction accuracy. Prop. \ref{pr:linfair} demonstrates that a fair regularizer must find the perpendicular parameter $w$ to $S_{*,1}$, however this constraint decreases classification accuracy. This study assumes a predetermined classifier. An alternative solution to fair recourse could be to incur a higher cost in finding the optimal action.
For example, if we add to the definition of the recourse problem that the optimal action should move instance $v$ and its twins to a favorable region, then individual fairness is embedded in the recourse definition. 

Now, we are trying to compute the adversarially robust recourse cost for $v$ and its twin.
Let $\Delta<1$ be perturbation radius and $B^{\mathcal{M}}_{\Delta \Plus}$ be its corresponding ACP.
To find the shape of ACP, the points $\mathbf{S}_\mathcal{M}^{\delta}(\mathbf{S}_\mathcal{M}^{-1}(v))$ should be calculated for each $\delta$ that belongs to $B_{\Delta}(0) = \{ \delta \in \mathcal{V}: \| \delta \|  \leq \Delta \}$. By additive intervention formula, $A$ satisfies the equation $A = f(V_{pa(A}) + U_A + \delta_A$ for every $\delta_A \in \mathbb{R}$ such that $-\Delta \leq \delta_A \leq \Delta$, therefore $A$ varies over continuous domains. 
As can be guessed, additive intervention is suitable for continuous features but difficult to apply to categorical variables in a meaningful way.
\paragraph{Challenge II}
\label{tx:cl2}
In the definition of $B^{\mathcal{M}}_{\Delta \Plus}(v)$  \cite{dominguez2022adversarial}, the authors assume that $\mathcal{V}$ is vector space with norm $\|.\|$, So the additive intervention is compatible with the vector space assumption.
While every discrete space is not a vector space, and every metric cannot necessarily be expressed with a norm. For example, 
let $d$ be the discrete metric:
\begin{equation}
\label{eq:dismetr}
d(x,y) = \begin{cases} 1, & x \neq y, \\ 0, & x = y. \end{cases}
\end{equation}
Then $d$ clearly does not satisfy the homogeneity property \ref{def:vspace} of the metric induced by a norm.
Hence, if we have a categorical variable with discrete space, we might not have a norm on space, so we need to use a metric instead of the norm to define ACP.
\paragraph{Challenge III} 
\label{tx:cl3}
Another problem in additive intervention is that the perturbation of categorical variable $A+\delta_A $ should be in $\mathcal{A}$, 
but it does not happen for every $\delta_A$ inside $B_{\Delta}$.
Additionally, the sum of nominal variable levels is meaningless. Thus, the additive intervention must be altered in categorical cases.

We can proceed to derive an explicit formula for the adversarial robust recourse cost.
As seen in Fig. ~\ref{fig:acp} (b), in linear SCMs the geometric shape of $B^{\mathcal{M}}_{\Delta \Plus}(0)$ is an ellipse (see Prop. \ref{pr:ellipse}). 
Consequently, the cost of a robust recourse is equivalent to the greatest distance of the ellipse from the decision boundary $L$. 
\begin{proposition} 
\label{pr:recCost}
For the recourse problem with conditions of Tab. \ref{table:condition}, the extra cost to have adversarially robust recourse for instance $v$ is:
\begin{equation}
\label{eq:robustCost}
r^{\mathcal{M}}_{\Delta \Plus}(v) - r^{\mathcal{M}}(v) = \dfrac{\Delta\|w^T_{|_X} \times \mathbf{S}^{\mathcal{M}}_{|_X} \|_{p^*}}{{\|w\|_{p^*}}}
\end{equation}
where $w_{|_X}$ and  $\mathbf{S}^{\mathcal{M}}_{|_X}$ are the restriction of $w$ and $\mathbf{S}^{\mathcal{M}}$ to continuous part $X$.
\end{proposition}
\paragraph{Challenge IV} 
\label{tx:cl4}
The propositions Prop. \ref{pr:linfair}
 and Prop. \ref{pr:recCost}
 demonstrate that in a linear scenario, individual fairness in recourse leads to individual fairness in adversarially robust recourse.
So as seen in Fig. \ref{fig:acp} (b), there is no individual fairness for adversarially robust recourse costs unless $w^T \bigcdot S_{*,1} = 0$.
If Instead of classical robust recourse \href{def:robrec}{problem},  try to find minimal action to move $\displaystyle \bigcup_{a \in \mathcal{A}} B^{\mathcal{M}}_{\Delta \Plus }(\ddot{v}_a)$ 
to favorable region,
then the adversarially robust recourse cost of each instance is fair w.r.t. the protected variable $A$ (see Fig. \ref{fig:acp} (c)).
%
%

\section{Fair Robust Recourse}
\label{sec:faro}
The previous chapter highlighted the challenges in obtaining fair and robust recourse. This section endeavors to define the recourse problem with fairness and robustness characteristics, and to formulate the definition of perturbations encompassing both continuous and categorical variables, as well as being unbiased w.r.t. protected variables.
We present a novel framework for defining an adversarially fair robust recourse problem and exhibit the existence of a solution with desirable properties. 
Furthermore, we demonstrate that in the new setting of the recourse problem, adversarial robustness implies individual fairness. Finally, we introduce the fair robust recourse (FARO) that is independent of the perturbation radius and possesses desirable properties.
\subsection{Counterfactual Perturbation}
\label{sec:perturbation}
\begin{figure}[ht]
\centering
\includegraphics[width =1\textwidth]{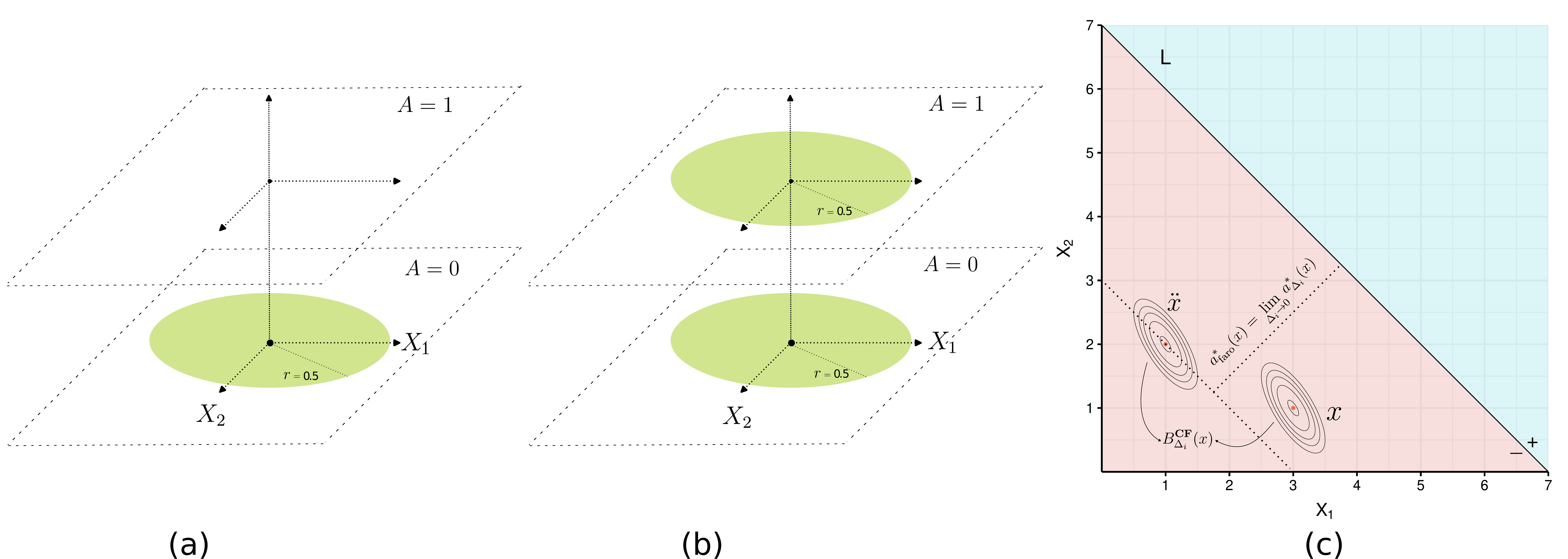}
 \caption{The shape of $B_{\frac{1}{2}}^{d}(0)$ with (a) discrete metric and (b) everywhere zero pseudometric for variable $A$.
 (c) FARO recourse is obtained by the limit of ADRR solution when the perturbation diameter $\Delta_i \rightarrow 0$.}
\label{fig:faro}
\end{figure}

As seen in Eq. \ref{eq:ACP}, the construction of ACP involves two steps. The first definition of the perturbation ball around $0$, and the second definition of the counterfactual perturbation by additive interventions.
By discussions of Challenge 2, the norm-based definition of perturbation does not cover all variable types, so we try to redefine counterfactual perturbation by means of metric instead of the norm.
Let $(\mathcal{X}_i, d_{\mathcal{X}_i})$ be a metric space correspond to the continuous variables $X_i$. For categorical variables, consider $(\mathcal{Z}_i, d_{\mathcal{Z}_i})$ as a pseudometric metric space (see App. \ref{def:pseudo} for more details).
Therefore, by product metric spaces (see App. \ref{def:prometes}), the space $(\mathcal{V} = \mathcal{Z} \times \mathcal{X},d)$  is also pseudometric space.
We can now define the perturbation ball after equipping space by metric.
\paragraph{Perturbation Ball.} 
For any instance $v$ in a (pseudo-)metric space $\mathcal{V}$ and a non-negative real number $\Delta$, the perturbation ball of radius $\Delta$ w.r.t the (pseudo-)metric $d$ is given by:
\begin{equation}
B_{\Delta}^{d}(v)=\{w \in \mathcal{V}:d(v,w) \leq \Delta\}.
\end{equation}
To complete the construction of ACP, we should compute $\mathbf{CF}(v, \delta; \mathcal{M})$ for every $\delta \in B_{\Delta}^{d}(v)$. 
The third Challenge states that additive intervention is not suitable for categorical variables, especially nominal types, as their corresponding metric space may not have an addition operator. 
Even if summability were not an issue, additive intervention provides no guarantee that the outcome value will remain in $\mathcal{Z}_i$ after disturbing the variable $\textbf{Z}_i$ and its parents by $\delta$. 
To solve these problems, we propose a middle intervention that combines 
\hyperref[eq:hard]{hard} intervention for categorical features and \hyperref[eq:soft]{soft} (additive) intervention for continuous parts. 
\begin{definition}[Middle Intervention]
\label{def:midint}
Consider SCM $\mathcal{M}$, with $n$ and $m$ continuous and categorical variables. 
Let the indexes of categorical  and continuous variables in vector $V$ be $\mathcal{I}_{\text{cat}}$  and $\mathcal{J}_{\text{con}}$ ($\mathcal{I}_{\text{cat}} \cup \mathcal{J}_{\text{con}} =\{1,2, \dots, n+m\}$).
The middle intervention $\mathcal{M}^{\theta_{\mathcal{I}}, \delta_{\mathcal{J}}}$  fixes the values of a subset $\mathcal{I} \subset \mathcal{I}_{\text{cat}}$ of categorical features $\mathbf{V}_{\mathcal{I}}$ to some fixed $\theta_{\mathcal{I}} \in\mathcal{Z}^{|\mathcal{I}|}$ and additive intervened the continuous features $\mathbf{V}_{\mathcal{J}}$ of a subset $\mathcal{J} \subset \mathcal{J}_{\text{con}}$ by some $\delta_{\mathcal{J}} \in \mathbb{R}^{|\mathcal{J}|}$ while preserving all other causal relationships.
The description of the structural equations for $\mathbb{S}_{\mathcal{M}}^{\delta_{\mathcal{I}},\theta_{\mathcal{J}}}$ follows:
\begin{equation}
\mathbf{V}_i := 
\begin{cases}
\theta_i & \text{if} \ i \in \mathcal{I} \\
f_i (\mathbf{V}_{\text{pa}(i)}, \mathbf{U}_i)+\delta_i, &  \text{if} \ i \in \mathcal{J}  \\
f_i (\mathbf{V}_{\text{pa}(i)}, \mathbf{U}_i) & \text{otherwise}
\end{cases}
\end{equation}
Similar to other interventions, the middle intervention's counterfactual is defined as 
$\mathbf{CF}\left(v, \theta_{\mathcal{I}}, \delta_{\mathcal{J}}; \mathcal{M}\right) = \mathbf{S}_{\mathcal{M}}^{\theta_{\mathcal{I}}, \delta_{\mathcal{J}}}(\mathbf{S}_{\mathcal{M}}^{-1}(v))$. 
\end{definition}
Now by using a middle intervention and perturbation ball, we are prepared to define a suitable counterfactual perturbation that encompasses categorical variables as well as continuous ones.
\begin{definition}[Counterfactual Perturbation]
\label{def:cp}
Let $\mathcal{M}$ be SCM satisfying the conditions specified in Def. \ref{def:midint}.
Consider a perturbation ball $B^{d_{\mathcal{I},\mathcal{J}}}_{\Delta}(v)$ for a subset of categorical and continuous variables with indices $\mathcal{I}$ and $\mathcal{J}$ and the induced metric $d_{\mathcal{I},\mathcal{J}}$ w.r.t. subspace $\mathcal{Z}^{|\mathcal{I}|}\times \mathcal{X}^{|\mathcal{J}|}$.
The counterfactual perturbation $B^{\mathbf{CF}}_{\Delta,\mathcal{I},\mathcal{J}}(v)$ for instance $v$ and perturbation radius $\Delta \in \mathbb{R}_{\geq 0}$ is the set of counterfactuals $v$ w.r.t middle interventions by the members of perturbation ball:
\begin{equation}
B^{\mathbf{CF}}_{\Delta,\mathcal{I},\mathcal{J}}(v) = \{ \mathbf{CF}\left(v, \theta_{\mathcal{I}}, \delta_{\mathcal{J}}; \mathcal{M}\right) \; : (\theta_{\mathcal{I}},\delta_{\mathcal{J}}+ v_{\mathcal{J}} ) \in B^{d_{\mathcal{I},\mathcal{J}}}_{\Delta}(v) \}
\end{equation}
where $\delta_{\mathcal{J}}$ is perturbation to change the value of  $v_{\mathcal{J}}$ to another instance that is inside in a perturbation ball $B^{d_{\mathcal{I},\mathcal{J}}}_{\Delta}(v)$.
\end{definition}
If all variables were intervened, indexes $\mathcal{I}$ and $\mathcal{J}$ could be dropped for a simpler notation $B^{\mathbf{CF}}_{\Delta}(v)$ and $\mathbf{CF}(v, \theta, \delta; \mathcal{M})$.
\begin{remark} \label{rm:zero}
	To modify the perturbation definition in \citet{dominguez2022adversarial} work that is suitable for categorical features, we use $B^{d_{\mathcal{I},\mathcal{J}}}_{\Delta}(v)$ instead of $B_{\Delta}(0)$ to define the perturbation shift.
 In this approach, we view $\delta_{\mathcal{J}}$ as a perturbation shift of continuous variables $V_{\mathcal{J}}$ and 
$\theta_\mathcal{I}$ as a categorical perturbation of variables $V_\mathcal{I}$ where $(\theta_{\mathcal{I}},\delta_{\mathcal{J}}+ v_{\mathcal{J}} ) \in B^{d_{\mathcal{I},\mathcal{J}}}_{\Delta}(v)$.
\end{remark}
The counterfactual perturbation can be decomposed into ACPs by the following proposition.
\begin{proposition}[Perturbation Decomposition]
\label{pr:ballShape}
Assume $B^{\mathbf{CF}}_{\Delta,\mathcal{I},\mathcal{J}}(v)$ is counterfactual perturbation with the conditions of Def. \ref{def:cp}.
Let $v = (z,x)$ be instance of $\mathcal{M}$ and 
$\Theta_{\Delta,\mathcal{I}} = \{ \theta \in \mathcal{Z}^{|\mathcal{I}|}: (\theta,.) \in B^{d_{\mathcal{I},\mathcal{J}}}_{\Delta}(v) \}$, the set of all categorical levels that inside the perturbation ball.
Then counterfactual perturbation can be decomposed as a union of ACPs:
\begin{equation}
\label{eq:union}
B^{\mathbf{CF}}_{\Delta,\mathcal{I},\mathcal{J}}(v) = \bigcup_{\theta \in \Theta_{\Delta,\mathcal{I}}} B^{\mathcal{M}^{do(\mathbf{V}_{\mathcal{I}} = \theta)}}_{\Delta_{\theta} \Plus}(\mathbf{CF}(v, \theta_{\mathcal{I}};\mathcal{M}))
\end{equation}
where $\Delta_{\theta}$ is continuous part value of $\Delta$. For example if the product metric is $L_2$ then $\Delta_{\theta} = \sqrt{\Delta^2- d_{\mathcal{Z}_\mathcal{I}}(\theta, z_{\mathcal{I}})^2}$.
\end{proposition}
Eq. \ref{eq:ACP} highlights the importance of the perturbation ball for determining optimal robust recourse. The form of the ball is influenced by the dissimilarity function $dist$, and if it is biased with respect to $A$, the recourse problem will also be unfair. To address this, all levels of $A$ must be treated equivalently with the chosen metric, which motivates the definition of protected features as a pseudometric space.
\begin{definition}[Protected Feature]
In $\mathcal{M}$ with categorical and continuous variables, consider a categorical variable $A \in \mathbf{Z}$ with pseudometric space $(\mathcal{A},d_{\mathcal{A}})$. 
The variable $A$ is called partially protected if there exist two levels such that their distance is $0$:  
\begin{equation}
\exists a,a' \in \mathcal{A} , \  \text{s.t.} \   d_{\mathcal{A}}(a, a') =0 \wedge a \neq a'
\end{equation}
If for all $a,a' \in \mathcal{A}$ we have $d_{\mathcal{A}}(a, a') =0$, then $A$ is called protected feature.
\end{definition}
The above definition ensures symmetry in counterfactual perturbation for small enough values of $\Delta$ (see Fig. \ref{fig:faro} (b)),
as described in the subsequent lemma.
\begin{lemma}[Perturbation Reduction Lemma] \label{pr:decompose}
If $A$ is protected and other categorical variables in $\mathcal{M}$ are not partially protected, then there exist $\Delta_0$ that for all $\Delta \leq \Delta_0$:
\begin{equation}
\label{eq:decompose}
B^{\mathbf{CF}}_{\Delta}(v) = \bigcup_{a \in \mathcal{A}} B^{\mathcal{M}^{do(A = a)}}_{\Delta \Plus}(\ddot{v}_a)
\end{equation}
as a result, for all $v, v' \in \ddot{\mathbbm{v}}$ we have $B^{\mathbf{CF}}_{\Delta}(v) = B^{\mathbf{CF}}_{\Delta}(v')$.
\end{lemma}
Lem.~\ref{pr:decompose} reduces categorical variables of $\mathcal{M}$ to the partially protected features, for solving the fair robust recourse problem. 
The lemma also shows that the set of twins and zero-radius counterfactual perturbations are equivalent.
\begin{corollary}
	\label{pr:twinDef}
	If $A$ is protected and other categorical variables in $\mathcal{M}$ are not partially protected, then the counterfactual twins equal zero-radius counterfactual perturbation
	$\ddot{\mathbbm{v}} = B^{\mathbf{CF}}_{0}(v) := \lim_{\Delta \rightarrow 0} B^{\mathbf{CF}}_{\Delta}(v)$.
\end{corollary}	
\subsection{Fair Robust Recourse}
We define an adversarially fair robust recourse problem by incorporating fairness properties through counterfactual perturbation $B^{\mathbf{CF}}_{\Delta}(v)$ that aligns with the protected group. 
\begin{definition}[Adversarially Fair Robust Recourse Problem (AFRR)]
\label{def:fairrobust}
Let $\mathcal{M}$ be SCM with conditions of Def. \ref{def:cp}. 
An adversarially fair robust recourse (AFRR) problem involves finding the minimum cost action that moves all $B^{\mathbf{CF}}_{\Delta, \mathcal{I}, \mathcal{J}}(v)$ into the favorable region:
\begin{equation} \label{def:affrp}
a^*_{\Delta}(v) =  
\underset{a \in\mathcal{F}(v)}{argmin}
 \; cost(v, a) \quad
\text{s.t.}\quad  h\left(\mathbf{CF}\left(v', a;  \mathcal{M}\right)\right)=1  \quad \forall v'
 \in B^{\mathbf{CF}}_{\Delta, \mathcal{I}, \mathcal{J}}(v)  
\end{equation}
\end{definition}
Similar to ~\citet{dominguez2022adversarial}, we say the action $a$ is \textbf{adversarially fair robust recourse} if satisfies the equation:
\begin{equation}
	    h(\mathbf{CF}(v', a;  \mathcal{M}))=1 \quad \forall v' \in B^{\mathbf{CF}}_{\Delta, \mathcal{I}, \mathcal{J}}(v)
\end{equation}
It is now possible to investigate the connection between the adversarially fair robust recourse action and the causal recourse problem.
\begin{proposition} 
\label{pr:optdec}
	Let $h(x) = \text{sign}(w \bigcdot v - b)$ be a linear classifier, $\mathcal{M}$ a linear SCM, and $B^{\mathbf{CF}}_{\Delta}(v)$ counterfactual perturbation.
	Then, the action $a$ is an adversarially fair robust recourse if and only for all $\theta \in \Theta_{\Delta}$, $\textit{a}$ is valid recourse action for the following modified classifiers:
	\begin{equation} \label{eq:subproblem}
	h_{\theta}(x) = \text{sign}(w \bigcdot v - (b + \Delta \| w^T \bigcdot S^a_{\theta} \|^*))
	\end{equation}
	where $\|\cdot\|^*$ denotes the dual norm of $\|\cdot\|$ and $S^a_{\theta}$ is denoted the mapping resulting from intervening by action $\textit{a}$ on structural equations of SCM $\mathcal{M}^{do(\mathbf{Z} = \theta)}$.
\end{proposition}
We now aim to demonstrate that the adversarially fair robust recourse problem is solvable for certain types of SCMs.
\begin{proposition}[Existence of AFRR Problem] 
\label{pr:faroEx}
	For a linear classifier $h$ and linear SCM
	$\mathcal{M}$, if there exists a continuous feature $X_j$ that
	is actionable and unbounded and $w^T \bigcdot {S_{\theta}}_{*,j} \neq  0$ for some $\theta \in \Theta_{\Delta}$ (i.e. $h$ depends on $X_j$ or its descendants), then there exists an adversarially fair robust recourse action for every $v \notin A_{\mathcal{M},h}^{\neq}$.
\end{proposition}
Similar to the definition of recourse cost, we define $r_{\Delta}^{\mathcal{M}}(v)$ for the AFRR problem. 
Lem.~\ref{pr:decompose} demonstrates that for small enough values of $\Delta$, $r_{\Delta}^{\mathcal{M}}(v) = r_{\Delta}^{\mathcal{M}}(v')$ holds for all $ v, v' \in \ddot{\mathbbm{v}}$. This fact highlights that AFRR implies individual fairness.
The recourse cost for the AFRR problem can be exactly computed with some mild conditions by combining Lem. \ref{pr:decompose} and \ref{pr:optdec} (see Fig. \ref{fig:chellenge} (c)).
\begin{corollary} \label{pr:indifair}
	Under Tab. \ref{table:condition}'s conditions, 
 	if $A$ is protected and other categorical variables in $\mathcal{M}$ are not partially protected, then there exist $\Delta_0$ that for $\Delta \leq \Delta_0$ and instance $v$ outside unfair area the AFRR problem has solution. The recourse cost of the AFRR problem is given by:
	\begin{equation}
	\begin{aligned}
	\label{eq:fairneed}
	r^{\mathcal{M}}_{\Delta}(v)=    \max_{a \in \mathcal{A}} \Big\{ \dfrac{|w^T\bigcdot \ddot{v}_a-b| + \Delta\|w^T.\mathbf{S}\|_{p^*}}{\|w\|_{p^*}}   \Big\}
	\end{aligned}
	\end{equation}
\end{corollary}
In the following proposition, we attempt to discover the relationship between the AFRR solution and individual fairness in the recourse problem.
\begin{proposition}
\label{pr:delta0}
For the SCM $\mathcal{M}$ with conditions of Lem. \ref{pr:decompose} we have:
\begin{enumerate}[label=(\alph*)]
	\item $\lim_{\Delta \rightarrow 0} r_{\Delta}^{\mathcal{M}}(v) = \max_{a \in \mathcal{A}} \{r^{\mathcal{M}}(\ddot{v}_a)\}$
	\item Recourse Problem is individually fair if and only if 
	$\displaystyle  \lim_{\Delta \rightarrow 0} r_{\Delta}^{\mathcal{M}}(v) = r^{\mathcal{M}}(v)$
\end{enumerate}
\end{proposition}
Prop. \ref{pr:delta0} has the idea to redefine algorithmic recourse in a way that is both fair and robust and independent of perturbation radius. We called this type of recourse \textbf{FARO} problem for abbreviation. 
\begin{definition}[FARO Recourse Problem]
	\label{def:fairrec}
	Assume $\big(\Delta_n\big)_{n = 1}^{\infty}$ be sequence of positive values such that $\Delta_n$ approaches $0$. Let the $a^*_{\Delta_n}$ be the optimal action according to the AFRR problem for counterfactual perturbation with radius $\Delta_n$. If the ${\displaystyle\lim_{n \rightarrow \infty} a^*_{\Delta_n}}$ exist,
 we say that the fair robust (FARO) recourse problem has a solution for instance $v$.
\end{definition}
The following result highlights the advantageous features of the newly defined recourse problem.
\begin{corollary}
	\label{pr:faroProp}
	The FARO recourse provides adversarial robustness and individual fairness simultaneously.
\end{corollary}
The existence of the FARO problem is similar to that of the AFRR problem. With its proof, we have completed all parts of the fair and robust recourse problem.
\begin{proposition}[Existence of FARO recourse]
	\label{pr:faroExist}
	For instance $v$, if the AFRR problem has a solution for some $\Delta$, then FARO recourse exists.
\end{proposition}
%
%
\section{Experiments}
\label{sec:experiments}
In this section, we validate our claims through experiments and assess the effects of different recourse definitions on individual fairness.
At first, we perform numerical simulations on various models and classifiers. Next, we apply our findings to both a real-world and semi-synthetic dataset as a case study.
The codes and instructions for reproducing our experiments are available at \href{https://github.com/Ehyaei/Fair-Robust-Algorithmic-Resource}{Github}.
\paragraph{Numerical Simulations}
Since the recourse actions require knowledge of the underlying SCM, we begin by defining two linear and non-linear ANM models for the SCM.
In our experiments, we utilize two non-protected continuous features ($X_i$) and a binary protected attribute ($A$) with a value of either 0 or 1. 
For each model, We generate $10,000$ samples by using the SCMs structural equation that is described in the App. \ref{app:simulation}.
To add the ground truth label, we consider both linear and non-linear functions in the form of $Y = sign(f(v,w) - b)$, where $w \in \mathbb{R}^{n+1}$ is coefficient of $V_i$ and $b \in \mathbb{R}$. We also examine an unaware baseline where $h$ does not depend on protected variable $A$.
For each dataset, we split the samples into 80\% for training and 20\% for testing. Then, we train a logistic regression (LR), support vector machine (SVM), and gradient boosting machine (GBM) using all features or just the non-protected features $\mathbf{X}$ as an unaware baseline.

We consider discrete and trivial pseudometric ($d(a,a')=0$ for all $a,a'$) for protected feature $A$. For continuous variables and product metric space, we use the $L_2$ norm.
The cost is defined as the $L_2$ norm, with $cost(v,a) = \|v - \mathbf{CF}(v,a) \|_2$.
Finally, we evaluate the methods presented in ~\cref{sec:faro} for having individual fairness by testing different perturbation radii $\Delta \in \{1, 0.5, 0.1\}$.
Since the main objective of this work is not to provide an algorithmic solution for causal recourse, we use a brute-force search to find the optimal action.
\paragraph{Case Studies}
We use the Adult Income Demographic dataset (ACSIncome) \cite{ding2021retiring}, an updated version of the UCI Adult dataset \cite{Dua:2019}, which contains over 195,000 records from California state in 2018. The data was obtained by using the Folktables Python package \cite{Folktables}. The data processing and modeling procedures adopted in this study are consistent with those reported in \citet{nabi2018fair} work. In addition, we analyze a semi-synthetic SCM proposed in \citet{karimi2020algorithmic} based on a loan approval scenario. For semi-synthetic data, all of the procedures are the same as in the numerical simulations section. See Appendix \ref{app:case_study} for further details.
\paragraph{Main Results}
We compare the results of our recourse problem by computing the recourse relative fairness ($\sigma_\textbf{R}$) by the formula: $\sigma_\textbf{R} = \displaystyle \frac{\max_{v \in \mathcal{V}} |r^{\mathcal{M}}(v)-r^{\mathcal{M}}(\ddot{v})|}{\frac{1}{|\mathcal{D}|}\sum_{v \in \mathcal{D}} r^{\mathcal{M}}(v)}$. 
We also consider the robust recourse relative fairness ($\sigma_\textbf{AR}$) and fair robust recourse relative fairness metrics ($\sigma_\textbf{FR}$) in a similar manner.
The simulation results are summarized in Tab. \ref{tab:sim_1}. The simulation confirms our expectations and shows that the fair robust recourse achieves individual fairness with $\sigma_\textbf{FR} = 0$.
\setlength{\extrarowheight}{-0.1cm}
\begin{table}[ht]
\small
\begin{tabular}{ c l  c c c  |  c c c || c c c | c c c }
	\toprule
	\centering
	\multirow{3}{*}{\textbf{Classifier}}
	& \multicolumn{6}{r}{\textbf{GT labels from \textit{linear}}} 
	& \multicolumn{6}{r}{\textbf{GT labels from \textit{nonlinear}}}
	\\
	\cmidrule(r){3-8} \cmidrule(r){9-14}
	& \multicolumn{3}{r}{\textbf{LIN}} 
	& \multicolumn{3}{r}{\textbf{ANM}}
	& \multicolumn{3}{r}{\textbf{LIN}}
	& \multicolumn{3}{r}{\textbf{ANM}} 
	\\
	\cmidrule(r){3-5} \cmidrule(r){6-8} \cmidrule(r){9-11} \cmidrule(r){12-14}
	&
	& $\sigma_\textbf{R}$
	& $\sigma_\textbf{AR}$
	& $\sigma_\textbf{FR}$
	& $\sigma_\textbf{R}$
	& $\sigma_\textbf{AR}$
	& $\sigma_\textbf{FR}$
	& $\sigma_\textbf{R}$
	& $\sigma_\textbf{AR}$
	& $\sigma_\textbf{FR}$
	& $\sigma_\textbf{R}$
	& $\sigma_\textbf{AR}$
	& $\sigma_\textbf{FR}$
	\\
	\midrule
	\multirow{6}{*}{\rotatebox[origin=c]{90}{Aware Label}} 
& GLM$(A,X)$ & 0.7 & 0.52 & \textbf{0.00} & 1.12 & 0.94 & \textbf{0.00} & 1.8 & 1.1 & \textbf{0.00} & 3.4 & 2.25 & \textbf{0.00} \\
& SVM$(A,X)$ & 0.79 & 0.57 & \textbf{0.00} & 1.16 & 0.95 & \textbf{0.00} & 0.99 & 0.62 & \textbf{0.00} & 2.01 & 1.44 & \textbf{0.00} \\
& GBM$(A,X)$ & 0.76 & 0.72 & \textbf{0.00} & 1.1 & 0.9 & \textbf{0.00} & 1.23 & 0.72 & \textbf{0.00} & 1.69 & 1.26 & \textbf{0.00} \\
& GLM$(X)$ & 0.62 & 0.43 & \textbf{0.00} & 1.22 & 0.99 & \textbf{0.00} & 1.36 & 0.83 & \textbf{0.00} & 2.29 & 1.62 & \textbf{0.00} \\
& SVM$(X)$ & 0.79 & 0.57 & \textbf{0.00} & 1.23 & 0.98 & \textbf{0.00} & 0.97 & 0.6 & \textbf{0.00} & 1.27 & 1 & \textbf{0.00} \\
& GBM$(X)$ & 0.79 & 0.58 & \textbf{0.00} & 1.19 & 0.96 & \textbf{0.00} & 1.03 & 0.62 & \textbf{0.00} & 1.44 & 1.1 & \textbf{0.00} \\
	\midrule
    \multirow{6}{*}{\rotatebox[origin=c]{90}{Unaware Label}} 
& GLM$(A,X)$ & 0.57 & 0.36 & \textbf{0.00} & 1.23 & 0.99 & \textbf{0.00} & 1.51 & 0.93 & \textbf{0.00} & 1.7 & 1.32 & \textbf{0.00} \\
& SVM$(A,X)$ & 0.6 & 0.39 & \textbf{0.00} & 1.14 & 0.92 & \textbf{0.00} & 0.88 & 0.56 & \textbf{0.00} & 2.09 & 1.5 & \textbf{0.00} \\
& GBM$(A,X)$ & 0.62 & 0.42 & \textbf{0.00} & 1.24 & 0.94 & \textbf{0.00} & 1.01 & 0.6 & \textbf{0.00} & 1.69 & 1.23 & \textbf{0.00} \\
& GLM$(X)$ & 0.55 & 0.36 & \textbf{0.00} & 1.25 & 1.01 & \textbf{0.00} & 1.37 & 0.83 & \textbf{0.00} & 2.38 & 1.7 & \textbf{0.00} \\
& SVM$(X)$ & 0.6 & 0.38 & \textbf{0.00} & 1.2 & 0.96 & \textbf{0.00} & 0.88 & 0.55 & \textbf{0.00} & 2.23 & 1.52 & \textbf{0.00} \\
& GBM$(X)$ & 0.62 & 0.75 & \textbf{0.00} & 1.25 & 0.95 & \textbf{0.00} & 0.98 & 0.59 & \textbf{0.00} & 1.83 & 1.32 & \textbf{0.00} \\
\\
	\hline
\end{tabular}
\caption{
The simulation results show a comparison of different classifiers with recourse relative fairness ($\sigma_\textbf{R}$), robust recourse relative fairness ($\sigma_\textbf{AR}$), and fair robust recourse relative fairness metrics ($\sigma_\textbf{FR}$) for $\Delta = 1$. The best methods for each dataset and metric are in \textbf{bold}, and only our fair robust methods achieved individual fairness in recourse.
}
\label{tab:sim_1}
\end{table}

\begin{figure}[ht]
\centering
\includegraphics[width =1\textwidth]{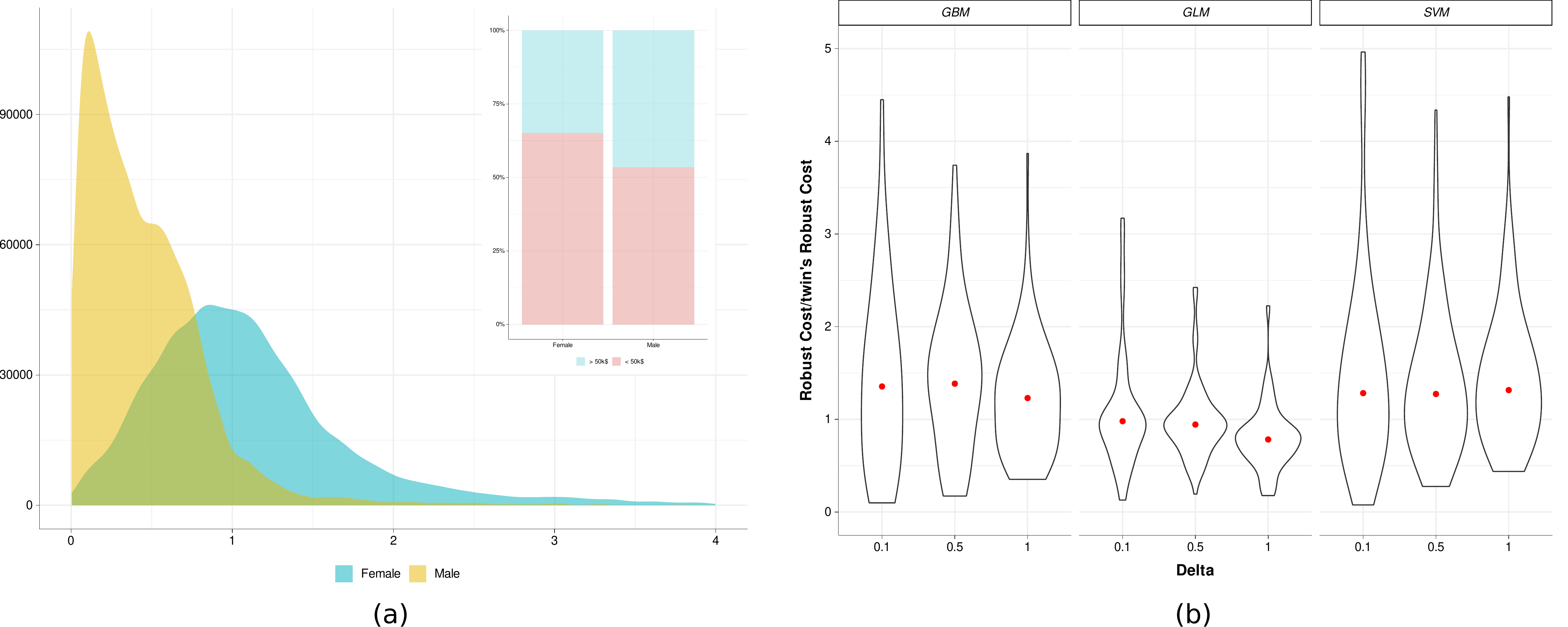}
 \caption{(a) The density plot of the ratio of recourse cost to its twin recourse cost by gender in ASCIncome indicates that females face a more difficult path to higher income, despite a similar proportion of males and females earning above 50k per year. (b) The Loan synthetic data's robust recourse cost ratio showed no fairness for any classifiers at any perturbation radius.
 }
\label{fig:data}
\end{figure}

In the ASCIncome dataset, we also look for individual fairness. This was accomplished by comparing the recourse cost's relative ratio to its twin recourse cost.
The results revealed that although the percentage of individuals earning above 50k per year between males and females is similar, females tend to face a harder path to achieving higher income compared to males (See Fig. \ref{fig:data} (a)).
In our study, we also compared the robust recourse cost ratio with its twin for the Loan synthetic data. The simulation results indicated that there is no fairness for any of the classifiers at any perturbation radius (See Fig. \ref{fig:data} (b)). On the other hand, the fair robust recourse demonstrated individual fairness for both datasets in the case study section.


\paragraph{Acknowledgement.}
We are grateful to Ricardo Dominguez-Olmedo for his insightful feedback.
The work of A. Ehyaei was supported by Grant 01IS20051 from the German Federal Ministry of Education and Research (BMBF).

\bibliography{refrences}  
\appendix


\section{Notation \& preliminaries}
\label{app:notation}
\begin{definition} 
\label{def:vspace}
	Given a vector space $X$ over a subfield $F$ of the complex numbers $\mathbb{C}$, a norm on $X$ is a real-valued function $p:X\to \mathbb {R}$ with the following properties, where $|s|$ denotes the usual absolute value of a scalar $s$:
	\begin{itemize}	
		\item Triangle inequality: $p(x+y)\leq p(x)+p(y)$ for all $x,y\in X.$
		\item Absolute homogeneity: $p(sx)=\left|s\right|p(x)$ for all $x\in X$ and all scalars $s$.
		\item Point-separating: for all $x\in X,$ if $p(x)=0$ then $x=0.$
	\end{itemize}
\end{definition}
\begin{definition}[Discrete metric Space]
Let $(X,d)$ be a metric space. (X,d) is called a discrete space if each $x \in X$ is an isolated point. In other words, there exists a $\delta >0$ such that for every $y \in X$ distinct from $x$ we have $d(x,y)>\delta$.
\end{definition}
\begin{definition}[pseudometric space] 
\label{def:pseudo}
A pseudometric space $(X,d)$ is a set $X$ together with a non-negative real-valued function $d:X\times X\longrightarrow \mathbb {R} _{\geq 0}$, called a pseudometric, such that for every $x,y,z \in X$,
\begin{itemize}
\item $d(x,x)=0.$
\item Symmetry: $d(x,y)=d(y,x)$
\item Triangle inequality: $d(x,z)\leq d(x,y)+d(y,z)$
\end{itemize}
The trivial example of pseudometric $d(x,y)=0$ for all $x,y\in X$
\end{definition}
\begin{definition}[Product Metric Spaces] 
\label{def:prometes}
If $(M_1,d_1)$, $\ldots$, $(M_n,d_n)$ are metric spaces, and $N$ is the norm on $\mathbb {R} ^{n}$, then ${\bigl (}M_{1}\times \cdots \times M_{n},d_{\times }{\bigr )}$ is a metric space, where the product metric is defined by:
\begin{equation*}
d_{\times }{\bigl (}(x_{1},\ldots ,x_{n}),(y_{1},\ldots ,y_{n}){\bigr )} =
N{\bigl (}d_{1}(x_{1},y_{1}),\ldots ,d_{n}(x_{n},y_{n}){\bigr )}
\end{equation*}
\end{definition}

\section{Proofs}
\label{sec:proof}

\begin{proposition} \label{pr:ellipse}
Let $\mathcal{M}$ be a linear SCMs with continuous manipulable features $X$, then $B^{\mathcal{M}}_{\Delta \Plus}(x_0)$ is ellipsoid with center of $x_0$.
\end{proposition}
\begin{proof}
By ellipsoid's definition, it is sufficient to show that $B^{\mathcal{M}}_{\Delta \Plus}(x)$, locus of points that are inside the equation $(x-x_0)^TA(x-x_0) = \Delta^2$, where $x_0$ is an ellipsoid's center and $A$ is a real, symmetric, positive-definite matrix. On other hand, $A$ is positive definite if and only if it decomposes to $M = B^TB$ with $B$ invertible \cite{horn2012matrix}. By the Eq. \ref{eq:ACPB},
$B^{\mathcal{M}}_{\Delta \Plus}(v) = \mathbf{S}_{\mathcal{M}}(B_{\Delta}(\mathbf{S}_{\mathcal{M}}^{-1}(x_0)))$. So if we consider $u_0 = \mathbf{S}_{\mathcal{M}}^{-1}(x_0)$, then $(U-u_0)^T \bigcdot (U-u_0) = \Delta^2$ by inverse mapping $(\mathbf{S}^{-1} \times (X-x_0))^T \bigcdot \mathbf{S}^{-1} \times (X-x_0) = \Delta^2$ therefore.
we have equation in $(X-x_0)^T \times (\mathbf{S}^{-1})^T \times \mathbf{S}^{-1} \times (X-x_0) = \Delta^2$. If consider
$A = (\mathbf{S}^{-1})^T \times \mathbf{S}^{-1}$ so by mentioned lemma it is positive-definite orthogonal decomposition property.
\end{proof}

\begin{proposition} \label{pr:levelsurf}
	The level curves of $L_p$ norms distance function of a point from hyperplane is also hyperplane.
\end{proposition}
\begin{proof}
	Let $L = \{x \in \mathbb{R}^n: w^T \bigcdot x  = b\}$ be the hyperplane where $w \in \mathbb{R}^n$ and $b \in \mathbb{R}$. 
	In the paper \cite{melachrinoudis1997analytical}, the minimum distance of point $x \in \mathbb{R}^n$ to hyperplane $L$ is calculated for arbitrary $L_p$ norm where $0 < p \leq \infty$, So we can write $dist(x,L) =  \frac{|w^T \bigcdot x - b|}{\|w\|_{p^*}}$. The level curves of the distance function are the locus of points that satisfies the equation $\{x \in \mathbb{R}^n: dist(x,L) = a\}$ for arbitrary $a \in \mathbb{R}$.
	\begin{equation*}
	dist(x,L) =  \dfrac{|w^T \bigcdot x - b|}{\|w\|_{p^*}} = a \quad \Leftrightarrow  \quad
	|w^T \bigcdot x - b| = a.\|w\|_{p^*} \quad \Leftrightarrow  \quad
	w^T \bigcdot x  = b \pm  a.\|w\|_{p^*}
	\end{equation*}
	Therefore, the level curves contain two hyperplanes that are located in both positive and negative space relative to hyperplane $L$.
\end{proof}	
\subsection{Proposition~\ref{pr:cfc}}
\begin{proof}
	\begin{enumerate}[label=(\alph*)]
		\item 
		Counterfactual additive intervention in ANMs can be thought of as adding $\delta$ to exogenous variables while keeping all structural equations unchanged, So we have:
		\begin{equation*}
		\mathbf{CF}(v,\delta;\mathcal{M}) =  \ \  S_{\mathcal{M}^{\delta}}(S^{-1}_{\mathcal{M}}(v)) = 
		 \ S \times (S^{-1} \times v + \delta) =  
		 S \times S^{-1}(v) + S \times \delta = 
		 v + S \times \delta 
		\end{equation*}
		\item 
		The hard intervention removes the effects of all parents of the variable $V_i$ and fixes its value by $\theta \in \mathbb{R}$. Therefore $S^{\theta}_{\mathcal{M}}$ is obtained by replace of $S^{\mathcal{M}}_{i,*}$ by $I_{i,*}$, in other words, $S_{\mathcal{M}}^{\theta} = S-S^0_{i} + I^0_{i}$, where $I$ is the identity $(n+1) \times (n+1)$ matrix and $M^0_i$ is the operator that converts a matrix $M$ into to matrix that all rows are equal to zero except the $i$-th row.
		To construct intervention mapping, we also need to fix the r.v. $U_i$ by constant value $\theta$ for each $u$. This work is done by the equation $S^{-1}(v) - (S^{-1}(v)_{i} - \theta)\times I_{i,*}$.
		So we can write:
		\begin{equation*}
		\begin{aligned}
		& \mathbf{CF}(v,do(V_i = \theta);\mathcal{M}) = 
		S_{\mathcal{M}}^{\theta}(S_{\mathcal{M}}^{-1}(v)) =
		(S-S^0_{i} + I^0_{i})\times (S^{-1}(v) - (S^{-1}(v)_{i} - \theta) \times I_{i,*}) = \\
		& S \times S^{-1}(v) - (S^{-1}(v)_{i} - \theta) *S  \times I_{i,*} +
		I^0_{i} \times S^{-1}(v) - S^0_{i} \times S^{-1}(v) +
		(S^{-1}(v)_{i} - \theta) * (S^0_{i}  \times I_{i,*} - I^0_{i} \times I_{i,*}) = \\
		& v + (\theta - S_{\mathcal{M}}^{-1}(v)_{i})S_{i,*}
		\end{aligned}
		\end{equation*}
		The last equation is true because, in additive noise models, we have $S[i,i] = S^{-1}[i,i] =1$, so this property implies $S^0_{i}  \times I_{i,*} = I^0_{i} \times I_{i,*}$ and $I^0_{i} \times S^{-1}(v) = S^0_{i} \times S^{-1}(v)$.
	\end{enumerate}
\end{proof}
\subsection{Proposition~\ref{pr:recdis}}
\begin{proof}
\begin{enumerate}[label=(\alph*)]
\item
By assumption $A$ is mutable, so for instance $v = (a,x)$ the counterfactual explanation has the same value for variable $A$. Therefore the optimal action only changes the continuous part. Since all continuous variables are manipulable, then optimal action can change all continuous values. Let $x^*$ be the continuous part of $\mathbf{CF}(v,a^*(v))$ corresponding to the optimal action $a^*(v) = do(V = \theta_{\mathcal{I}})$, then $x^*$ is also the solution of below optimization problem:
\begin{equation}\label{eq:destoLine}
\min_{x' \in \mathbb{R}^n} \|(a,x)-(a,x') \|_{p} \quad \text{s.t.} \quad h((a,\theta)) = 1 \quad \Leftrightarrow \quad
\min_{x' \in \mathbb{R}^n} \|x-x'\|_{p} \quad \text{s.t.} \quad \sum_{i = 1}^{n} w_i x_i' - b + w_0a = 0
\end{equation}
The solution of Eq. \ref{eq:destoLine} is equivalent to finding the minimum distance of continuous part $v$ to the line $L$ with the equation 
$L = \{y \in \mathbb{R}^n: \sum_{i = 1}^{n} w_i y_i - b + w_0a = 0 \}$.
In the work \citet{melachrinoudis1997analytical}, the minimum distance of the point to hyperplane is calculated for arbitrary $L_p$ norm where $0 < p \leq \infty$, So by using its result, we can write for the optimal value of $x^*$:
\begin{equation*}
\| x- x^* \|_p =  \dfrac{|\sum_{i = 1}^{n} w_i x_i - b + w_0a|}{\|w\|_{p^*}} = 
\dfrac{|w^T \bigcdot v-b|}{\|w\|_{p^*}}  = r^{\mathcal{M}}(v)
\end{equation*}
\item
If $A$ is mutable, its value is modified in order to find the best recourse of action.
Therefore the corresponding counterfactual of optimal action $\mathbf{CF}(v,a^*(v)) = (a^*,x^*)$  satisfies in the below optimization problem: 
\begin{equation*}
\begin{aligned}
& r^{\mathcal{M}}(v)  = \min_{a' \in \mathcal{A}, x' \in \mathbb{R}^n} \big\{ \|(a,x)-(a',x') \|_{p} \big \} \quad \text{s.t.} \quad h((a',x')) = 1 \Leftrightarrow \\
& \min_{a' \in \mathcal{A}} \{ \min_{x' \in \mathbb{R}^n} \{\|(a,x)-(a',x') \|_{p}\} \} \quad \text{s.t.} \quad h((a',x')) = 1  \Leftrightarrow \\
& \min_{a' \in \mathcal{A}} \{|a-a'|^p + \min_{x' \in \mathbb{R}^n} \{\|x-x'\|^p_{p}\} \} \quad \text{s.t.} \quad h((a',x')) = 1  \\
\end{aligned}
\end{equation*}
So when we fix the value $A = a'$, like as part (a), the optimization problem is equivalent to finding the minimum distance of continuous part $v$ to the line: 
\begin{equation*}
L_{a'} = \{y \in \mathbb{R}^n: \sum_{i = 1}^{n} w_i y_i - b + w_0a' = 0 \}
\end{equation*}
By the last equation and the distance equation in section (a) we can write: 
\begin{equation*}
r^{\mathcal{M}}(v)  = \min_{a' \in \mathcal{A}} \bigg\{\big(|a-a'|^p +  \dfrac{|w_0 a' +\sum_{i = 1}^{n} w_i x_i -b|^p}{\|w\|_{p^*}^p}\big)^{\frac{1}{p}} \bigg\}
\end{equation*}

\item When $A$ is immutable the counterfactual point $\mathbf{CF}(v,a^*(v)) = (a,x^*)$ according additive or hard action lies on the line $L$ with equation
$L = \{y \in \mathbb{R}^n: \sum_{i = 1}^{n} w_i y_i - b + w_0a = 0 \}$.
By the result of section 5  of \citet{melachrinoudis1997analytical}'s work, we can write the coordinates of $x^*$ as below formula:
\begin{equation*}
\begin{cases}
v_i - \dfrac{|w^T \bigcdot v-b|. |w_i|^{\frac{1}{p-1}}.sign(\frac{w^T \bigcdot v-b}{w_i})}{\|w\|_{p^*}}  & w_i \neq 0 \\
v_i  & w_i = 0 \\
\end{cases}
\end{equation*}
Therefore $a_{\theta}^*(v) = do(V = v + (0, \eta))$. If we use additive intervention by Prop. \ref{pr:cfc} $v + (0, \eta) = \mathbf{CF}(v,\delta;\mathcal{M}) = v + S\times \delta$, so we have $\delta = (0, \eta) \times (S^{-1})^T$.

In the case where $A$ is mutable, let $a^*$ be maximum levels $\mathcal{A}$ that satisfies the Eq. \ref{eq:mrecdis}. Then set $\mathbf{CF}(v,a^*(v)) = (a^*,x^*)$, so the point $x^*$ lies on the line 
$L_{a^*} = \{y \in \mathbb{R}^n: \sum_{i = 1}^{n} w_i y_i - b + w_0a^* = 0 \}$.
So for $w_i \neq 0$ if we define $\gamma_{a^*} = {\sum_{i  = 1}^{n}} w_i.x_i -b + w_0.a^*$ we have:
\begin{equation*}
x^*_i = v_i - \dfrac{|\gamma_{a^*}|. |w_i|^{\frac{1}{p-1}}.sign(\frac{\gamma_{a^*}}{w_i})}{\|w\|_{p^*}} = 
v_i - \dfrac{|\tiny{w^T \bigcdot v-b'}|. |w_i|^{\frac{1}{p-1}}.sign(\frac{w^T \bigcdot v-b'}{w_i})}{\|w\|_{p^*}} 
\end{equation*}
where $b' = b-w_0.(a-a')$, so the last equation completes the proof.
\item
When $A$ is mutable, Let $v = (a, x)$ be an arbitrary instance and $(a, x_i^*)$ is corresponding optimal counterfactual that lies on the boundary line $L$ for each norm $\|.\|_{p_i}$, Therefore by section (a) of the Prop.~\ref{pr:recdis}, the $r_{i}^{\mathcal{M}}(v) = \|x-x_i^*\|_{p_i} = \frac{|w^T \bigcdot v-b|}{\|w\|_{p_i^*}}$ for each $i$. Therefore we can write for each arbitrary point $(a,x')$ on line $L$:
\begin{equation*}
	\|x-x'\| =  \sum_{i = 1}^{n} \alpha_i \|x-x'\|_{p_i} \geq
	\sum_{i = 1}^{n} \alpha_i \|x-x^*_{i}\|_{p_i}  = \sum_{i = 1}^{n} \alpha_i\dfrac{|w^T \bigcdot v-b|}{\|w\|_{p_i^*}} =
	\sum_{i = 1}^{n} \alpha_i  r_{i}^{\mathcal{M}}(v)
\end{equation*}
where the last inequality follows by the fact that for each $i$ we have $\|x-x'\|_{p_i} -\frac{|w^T \bigcdot v-b|}{\|w\|_{p_i^*}} \geq 0$ because $x^*_i$ has the shortest distance from the line. Then for $(a.x')$ lies on the line $L$ we have:
\begin{equation*}
r^{\mathcal{M}}(v) = \min_{x' \in \mathbb{R}^n} \|(a,x)-(a,x') \| = 
\min_{x' \in \mathbb{R}^n} \|x-x'\| \geq \sum_{i = 1}^{n} \alpha_i  r_{i}^{\mathcal{M}}(v)
\end{equation*}
To prove another side of the inequality, we use the fact that the norms $L_p$ are decreasing w.r.t $p$ (is proved by direct use of Holder's inequality). So for each $i$ we have $\|.\|_{p_i} \leq \|.\|_{p_{i_*}}$, consequently:
\begin{equation}
\|x-x'\| = \sum_{i}^{} \alpha_i \|x-x'\|_{p_i} \leq
 \sum_{i}^{} \alpha_i \|x-x'\|_{p_{i_*}} = \|x-x'\|_{p_{i_*}}
\end{equation}
Therefore the minimum of $\|x-x'\|$ is less than minimum value of 
$\|x-x'\|_{p_{i_*}}$, this fact results another side of inequality. 
The proof in the case where $A$ is mutable is similar, so we omit its equations.
\end{enumerate}
\end{proof}
\subsection{Corollary~\ref{pr: twins}}
\begin{proof}
By assumption $\mathcal{M}$ has only one categorical variable, Since $\mathcal{M}$ is linear SCM, so we can assume that $A$ is not influenced by any continuous variable and therefore has no parents.
In this setting by additive noise model, we have $A = U_A$. So the intervention of $\text{do}(A = a')$ is equal to set random variable $U_A$ as constant $U_A = a'$.
Without loss of generality we can suppose $V_1 = A$, so $S_{1,*}^T =  (1,0,\dots,0)$. It results 
$S^{-1}(v)_1 = a$.
Finally by the Eq.~\ref{eq:hardInterv}, for computing counterfactual twins of $v = (a,x)$ respect the level $a'$, we can write:
\begin{equation*}
\ddot{v}_{a'} = v + (a'-S^{-1}(v)_1) . S_{*,1} =
v +(a' - a) . S_{*,1} 
\end{equation*}
where $S_{*,1}$ is the first column of the matrix $\mathbf{S}_{\mathcal{M}}$.
\end{proof}
\subsection{Proposition~\ref{pr:unfairEq}}
\begin{proof}
At first, we show that each instance $v = (a,x)$ in the unfair area of \ref{pr:unfairEq} is individually unfair w.r.t classifier $h$. We can suppose that for the levels of $A$ we have $a_1 < a_2 < \dots <a_k$. 
Without loss of generality suppose $v = (a_1,x)$. If $v$ is individually fair, so for $v$ and its twin $\ddot{v}_{a_k} = v+(a_k - a_1) . S_{*,1}$ we must have $w^T \bigcdot v \leq b$ and $w^T \bigcdot \ddot{v}_{a_k} \leq b$. In other hand we can suppose $w^T \bigcdot S_{*,1} >0$. It means the direction of vector $S_{*,1}$ is in the same direction as vector $w$. It results that the point $w^T \bigcdot \ddot{v}_{a_k} = w^T \bigcdot v + (a_k - a_1) . w^T \bigcdot S_{*,1}$ is closer to $L$ respect to $v$. Since the distance between $v$ and  $\ddot{v}_{a_k}$ in direction $w$ is more that the diameter of the unfair region, so if $v$ in $P^{-}$ then $\ddot{v}_{a_k}$ must be in the other side of $L$ i.e. in $P^{+}$ and it is contradicted by assumption.

Conversely, suppose $v$ is individually unfair, so there exists $i$ s.t. $h(v) \neq h(\ddot{v}_{a_i})$. Since $a_k > a_i$ and $w^T \bigcdot S_{*,1} >0$ then $h(v)\neq h(\ddot{v}_{a_k})$. It means that $v$ and $\ddot{v}_{a_k}$ are in different side of L. Therefore the distance $v$ from $L$ is less than distance between $v$ and $\ddot{v}_{a_k}$. It means $dist(v,L) \leq dist(v, \ddot{v}_{a_k})$ that can be in different classes. It completes the proof.  
\end{proof}
\subsection{Proposition~\ref{pr:linfair}}
\begin{proof}
When $w^T \bigcdot S_{*,1} = 0$, Prop.~\ref{pr:unfairEq} causes that classifier $h$ is individually fair for any instance of $v$. Conversely if classifier $h$ is individually fair, then for all $a' \in \mathcal{A}$ by the Eq. \ref{eq: twins} and \ref{eq:recdis} we have:
\begin{equation*}
\begin{aligned}
& r^{\mathcal{M}}(\ddot{v}_{a'}) =  \dfrac{|w^T \bigcdot (v+ (a' - a) \times S_{*,1}) - b|}{\|w\|_{p^*}} = 
\dfrac{|w^T \bigcdot v + (a' - a) w^T \bigcdot S_{*,1} -b|}{\|w\|_{p^*}} = r^{\mathcal{M}}(v) \Rightarrow \\  
& |w^T \bigcdot v + (a' - a) w^T \bigcdot S_{*,1}-b| = \text{constant} \quad \forall a' \in \mathcal{A} \Rightarrow (a' - a) w^T \bigcdot S_{*,1} = 0.
\end{aligned}
\end{equation*}
The proof is completed by the final equation.
\end{proof}
\subsection{Proposition~\ref{pr:recCost}}
\begin{proof}
At first, we can suppose that the values of $A$ are fixed inside the $B^{\mathcal{M}}_{\Delta \Plus}(v)$. It happens when $\Delta \leq \min_{a,a'}| a-a'|$.
If $p = 2$, by the Prop. \ref{pr:ellipse}, the shape of $B^{\mathcal{M}}_{\Delta \Plus}(v)$ is ellipsoid with center of $v$, so in general case, we can consider $B^{\mathcal{M}}_{\Delta \Plus}(v)$ as ellipsoid in $L_p$ norm (convex region respect to norm).
By the Prop. \ref{pr:levelsurf}, in $L_p$ norm the level surfaces of distance function point from boundary hyperplane $L$ are also hyperplanes which are parallel to plane $L$. So we can write the distance of the ellipsoid to plane $L$ as the maximum distance of the ellipsoid to its center at the direction of $w$ that is denoted by $d_{\text{max}}$ in addition to the minimum distance of $v$ from plane $L$ that is equal $r^{\mathcal{M}}(v)$.

Without loss of generality suppose $v$ is on the origin. 
Since the $B^{\mathcal{M}}_{\Delta \Plus}(v)$ has only continuous part perturbation, then $d_{\text{max}}$ is the minimum distance of continuous part from the origin in the direction $w_{|_X}$. Therefore we can suppose that $\mathcal{M}$ has only continuous features with mapping $\mathbf{S}^{\mathcal{M}}_{|_X}$ and linear classifier $h$ has coefficients $w^T_{|_X}$. 
Set objective function $f(v) = \frac{|w^T \bigcdot v|}{\|w\|_{p^*}}$ subject to  $g(v) = \|S^{-1}v\|_p = \Delta$. To find the minimum of $f$ we use the Lagrange multiplier. 
To compute gradient $\nabla g$, if consider $h(v) = \| v\|_p$, so $g(v) = h(S^{-1}v)$. By the gradient chain rule, we have
$\nabla g (v)= (S^{-1})^T\nabla h(S^{-1}v)$ \cite{petersen2008matrix}. On other hand,
the gradient $\nabla h$ is obtained by partial derivation of $L_p$ norm. So for $j=1,2, \dots ,n$, by chain rule, we have:
\begin{equation}
\partial_j \|\mathbf{v}\|_p = \frac{1}{p} \left(\sum_i \vert v_i \vert^p\right)^{\frac{1}{p}-1} \cdot p \vert v_j \vert^{p-1} \text{sign}(v_j) =  \left(\frac{\vert v_j \vert}{\|\mathbf{v}\|_p}\right)^{p-1} \text{sign}(v_j)
\end{equation}

Since $B^{\mathcal{M}}_{\Delta \Plus}(x)$ is symmetric around its center then we can suppose that $f(x) = \frac{w^T \bigcdot x}{\|w\|_{p^*}}$. 
Let $u = S^{-1}(v)$, be value of inverse mapping of $v$. 
The optimum point By the Lagrange multiplier satisfies the equation:
\begin{equation}\label{eq:grad}
\nabla f\,\|\,\nabla g \quad \Leftrightarrow \quad 
(S^{-1})^T\nabla h(S^{-1}(v)) \,\|\, w \quad \Leftrightarrow \quad 
(|u_i|^{p-1}\text{sign}(u_i))_{i=1}^n = \lambda (S^Tw)^T
\end{equation}
Let $(c_i)_{i=1}^n $ be the values of vector $(S^Tw)^T$. By this notation, we can write Eq. \ref{eq:grad} in the simpler version: 
\begin{equation} \label{eq:simple}
	|u_i| = |\lambda|^{\frac{1}{p-1}}.|c_i|^{\frac{1}{p-1}}
\end{equation}
To find $\lambda$ we put the Eq. \ref{eq:simple} in the subjective function:
\begin{equation}
	\sum_{i = 1}^n |u_i|^p = \Delta^p \Leftrightarrow \quad 
	\sum_{i = 1}^n |\lambda|^{\frac{p}{p-1}}.|c_i|^{\frac{p}{p-1}} = \Delta^p \Leftrightarrow \quad 
	|\lambda|^{p^*}.\| S^Tw\|_{p^*}^{p^*} = \Delta^p \Leftrightarrow \quad  
	\lambda = \dfrac{\Delta^{p-1}}{\| S^Tw\|_{p^*}}
\end{equation}
Finally, put $v$ in objective function $f$ to find optimum distance:
\begin{equation}
\begin{aligned}
 f(v) = & \frac{|w^T \bigcdot v|}{\|w\|_{p^*}} =  \frac{1}{\|w\|_{p^*}}|w^T S \times u|  =  \frac{1}{\|w\|_{p^*}}|u^T \bigcdot S^Tw| = 
\frac{1}{\|w\|_{p^*}} \sum_{i  = 1}^{n} |c_i| |\lambda|^{\frac{1}{p-1}}.|c_i|^{\frac{1}{p-1}} = \\
& \frac{|\lambda|^{\frac{1}{p-1}}}{\|w\|_{p^*}} \sum_{i = 1}^{n} |c_i|^{\frac{p}{p-1}} = 
\frac{\Delta}{\| S^Tw\|_{p^*}^{\frac{1}{p-1}} \|w\|_{p^*}} \| S^Tw\|_{p^*}^{\frac{p}{p-1}} = \Delta \dfrac{\| S^T \times w\|_{p^*}}{\|w\|_{p^*}}
\end{aligned}
\end{equation}
The last equation completes the proof.
\end{proof}
\subsection{Proposition~\ref{pr:ballShape}}
\begin{proof}
Writing the definitions yields the proof directly.
By the Def. \ref{def:cp}, the $B^{\mathbf{CF}}_{\Delta,\mathcal{I},\mathcal{J}}(v)$ is equal to:
\begin{equation*}
\begin{aligned}
& B^{\mathbf{CF}}_{\Delta,\mathcal{I},\mathcal{J}}(v) =  
\{ \mathbf{CF}\left(v, \theta_{\mathcal{I}}, \delta_{\mathcal{J}}; \mathcal{M}\right) \; : (\theta_{\mathcal{I}},\delta_{\mathcal{J}}+ v_{\mathcal{J}} ) \in B^{d_{\mathcal{I},\mathcal{J}}}_{\Delta}(v) \} = \\
& \bigcup_{\theta \in \Theta_{\Delta,\mathcal{I}}} \{ \mathbf{CF}\left(v, \theta, \delta_{\mathcal{J}}; \mathcal{M}\right) \; : (\theta,\delta_{\mathcal{J}}+ v_{\mathcal{J}} ) \in B^{d_{\mathcal{I},\mathcal{J}}}_{\Delta}(v) \} = \\
& \bigcup_{\theta \in \Theta_{\Delta,\mathcal{I}}} \{ \mathbf{CF}\left(v, \theta, \delta_{\mathcal{J}}; \mathcal{M}\right) \; : (\theta,\delta_{\mathcal{J}} ) \quad \text{s.t} \quad 
d_{\times}(d_{\mathcal{Z}_\mathcal{I}}(\theta, z_{\mathcal{I}}), d_{\mathcal{X}_\mathcal{J}}(\delta_{\mathcal{J}} + v_{\mathcal{J}}, v_{\mathcal{J}})) \leq \Delta \} = \\ 
& \bigcup_{\theta \in \Theta_{\Delta,\mathcal{I}}} \{ \mathbf{CF}\left(v, \theta, \delta_{\mathcal{J}}; \mathcal{M}\right) \; : \delta_{\mathcal{J}} \in  B_{\Delta_{\theta}}(0_{\mathcal{I}})\} = 
\bigcup_{\theta \in \Theta_{\Delta,\mathcal{I}}} \{ \mathbf{CF}\left(v, \delta_{\mathcal{J}}; \mathcal{M}^{do(\mathbf{V}_{\mathcal{I}} = \theta)}\right) \; : \delta_{\mathcal{J}} \in  B_{\Delta_{\theta}}(0_{\mathcal{I}})\} = \\
& \bigcup_{\theta \in \Theta_{\Delta,\mathcal{I}}} B^{\mathcal{M}^{do(\mathbf{V}_{\mathcal{I}} = \theta)}}_{\Delta_{\theta} \Plus}(\mathbf{CF}(v, \theta_{\mathcal{I}};\mathcal{M}))
\end{aligned}
\end{equation*}
The proof is completed by the last equation, where $d_{\times}$ is the product norm.
\end{proof}
\subsection{Lemma~\ref{pr:decompose}}
\begin{proof}
Let $v = (z,x)$ be an instance of $\mathcal{M}$ and $\mathcal{A} = \{a_1, a_2, \dots, a_k \}$ levels of $A$. Since $A$ is a protected feature and other categorical variables are not partially protected, and given that categorical variables have a discrete topology, then there exist $\Delta_0$ such that for $\Delta \leq \Delta_0$ the set
$\Theta_{\Delta} = \{ z' \in \mathcal{Z}: (z',.) \in B^{d}_{\Delta}(v)\}$ contains $z'$-values that all categorical variables except $A$ have fixed value (see Fig. \ref{fig:faro}(a)). To find $\Delta_0$ It is sufficient to consider
\begin{equation*}
\Delta_0 = \min_{z,z' \in \mathcal{Z}}d_{\mathcal{Z}}(z,z') \quad \text{s.t.} \quad \Delta_0 > 0
\end{equation*}
Since $\mathcal{Z}$ is finite set, the $\Delta_0$ exists.
Thus we can suppose that $\mathcal{M}$ has only one categorical variable $A$. It results
$\Theta_{\Delta} = \{a_1,a_2, \dots, a_k \}$. 
On the other hand, since A is a protected group $d_{\mathcal{A}}(a,a') = 0$, then $\Delta_{\theta} = \Delta$.
By use the Proposition~\ref{pr:ballShape} we can write:
\begin{equation*}
\begin{aligned}
B^{\mathbf{CF}}_{\Delta}(v)  =  \bigcup_{\theta \in \Theta_{\Delta}} B^{\mathcal{M}^{do(\mathbf{Z}= \theta)}}_{\Delta_{\theta} \Plus}(\mathbf{CF}(v, \theta;\mathcal{M})) = 
\bigcup_{a \in \mathcal{A}} B^{\mathcal{M}^{do(A= a)}}_{\Delta \Plus}(\mathbf{CF}(v, a;\mathcal{M})) = 
\bigcup_{a \in \mathcal{A}} B^{\mathcal{M}^{do(A= a)}}_{\Delta \Plus}(\ddot{v}_a) 
\end{aligned}
\end{equation*}
The last equation completes the proof.
\end{proof}
\subsection{Corollary~\ref{pr:twinDef}}
\begin{proof}
By using the Lemma~\ref{pr:decompose} we can write:
\begin{equation*}
B^{\mathbf{CF}}_{0}(v)  = \lim_{\Delta \rightarrow 0} B^{\mathbf{CF}}_{\Delta}(v) = 
\lim_{\Delta \rightarrow 0} \bigcup_{a \in \mathcal{A}} B^{\mathcal{M}^{do(A= a)}}_{\Delta \Plus}(\ddot{v}_a) = 
\bigcup_{a \in \mathcal{A}} \ddot{v}_a = \ddot{\mathbbm{v}} 
\end{equation*}
We can suppose the structural equations respect additive intervention is continuous, So the last equation is true because each ball $B^{\mathcal{M}^{do(A= a)}}_{\Delta \Plus}(\ddot{v}_a)$ is around $\ddot{v}_a$ so when it shrinks continuously then its limit goes to $\ddot{v}_a$.
\end{proof}
\subsection{Proposition~\ref{pr:optdec}}
\begin{proof}
	The proof is similar to the proof of Proposition 3 of the work \citet{dominguez2022adversarial}. 
	By the definition of adversarially fair robust recourse action, the $a(\pi_{\mathcal{I}}, \tau_{\mathcal{J}})$ satisfies the equation:
	\begin{equation*}
	h(\mathbf{CF}(v', a;  \mathcal{M}))=1 \quad \forall v' \in B^{\mathbf{CF}}_{\Delta}(v) 
	\end{equation*}
	By Prop. \ref{pr:ballShape} we can replace $B^{\mathbf{CF}}_{\Delta}(v)$ by $\bigcup_{\theta \in \Theta_{\Delta}} B^{\mathcal{M}^{do(\mathbf{Z} = \theta)}}_{\Delta_{\theta} \Plus}(\mathbf{CF}(v, \theta;\mathcal{M}))$,
	So for every $\theta \in \Theta_{\Delta}$, the action $a$ satisfies in optimization problem:
	\begin{equation*}
	\forall v' \in B^{\mathcal{M}^{do(\mathbf{V} = \theta)}}_{\Delta_{\theta} \Plus}(\mathbf{CF}(v, \theta;\mathcal{M})) \quad \quad
	h(\mathbf{CF}(v', a;  \mathcal{M}^{do(\mathbf{V}_{\mathcal{I}} = \theta_{\mathcal{I}})}))=1 
	\end{equation*}
	For simplicity we use $\mathcal{M}^{\theta}$ instead of 
	$\mathcal{M}^{do(\mathbf{Z} = \theta)}$ , and $v_{\theta}$ instead of $\mathbf{CF}(v, \theta;  \mathcal{M})$. Then for every $\theta \in \Theta_{\Delta}$ we have:
	\begin{equation} \label{eq:opt}
	h(\mathbf{CF}( v', a;  \mathcal{M}^{\theta})) =1 \quad
	\forall v' \in B^{\mathcal{M}^{\theta}}_{\Delta_{\theta} \Plus}(v_{\theta}) \quad \Leftrightarrow \quad 
	\min_{v' \in B^{\mathcal{M}^{\theta}}_{\Delta_{\theta} \Plus}(v_{\theta})} w^T \bigcdot \mathbf{CF}(v', a;  \mathcal{M}^{\theta}) \geq b
	\end{equation}
	We can write each element $v'$ in $B^{\mathcal{M}^{\theta}}_{\Delta_{\theta} \Plus}(v_{\theta})$ by $v' = \mathbf{CF}(v, \eta;  \mathcal{M}^{\theta})$ for some $\|\eta\| \leq \Delta_{\theta}$.	
	Under the assumption that the $\mathcal{M}$ is linear, the intervention map $\mathbf{S}^{a}_{\theta}$ of action $a$ on 
	structural equation $\mathbb{S}_{\mathcal{M}^{\theta}}$ is also linear, so we can write:
	\begin{equation}
	\begin{aligned}
	& \mathbf{CF}(v', a;\mathcal{M}^{\theta}) = \mathbf{S}^{a}_{\theta}(\mathbf{S}_{\mathcal{M}^{\theta}}^{-1}(v'))
	=\mathbf{S}^{a}_{\theta}(\mathbf{S}_{\mathcal{M}^{\theta}}^{-1}(\mathbf{CF}(v, \eta;  \mathcal{M}^{\theta})))
	=\mathbf{S}^{a}_{\theta}(\mathbf{S}_{\mathcal{M}^{\theta}}^{-1} (\mathbf{S}_{\mathcal{M}^{\theta}}^{\eta}(\mathbf{S}_{\mathcal{M}^{\theta}}^{-1}(v)))) = \\ 
	&\mathbf{S}^{a}_{\theta}(\mathbf{S}_{\mathcal{M}^{\theta}}^{-1}(\mathbf{S}_{\mathcal{M}^{\theta}}(\mathbf{S}_{\mathcal{M}^{\theta}}^{-1}(v)+\eta))) 
	= \mathbf{S}^{a}_{\theta}(\mathbf{S}_{\mathcal{M}^{\theta}}^{-1}(v)+\eta)
	= \mathbf{S}^{a}_{\theta}(\mathbf{S}_{\mathcal{M}^{\theta}}^{-1}(v)) +\mathbf{S}^{a}_{\theta}(\eta)
	= \mathbf{CF}(v, a;\mathcal{M}^{\theta}) +S^{a}_{\theta}\eta
	\end{aligned}
	\end{equation}
	where $S^{a}_{\theta}$ denotes the inter-space mapping resulting from hard-intervention $a$ on the $\mathcal{M}^{\theta}$.
	Then for the ACP $B^{\mathcal{M}^{\theta}}_{\Delta_{\theta} \Plus}(v_{\theta})$, the Eq. \ref{eq:opt} is equivalent to:
	\begin{equation}
    \begin{aligned}
	& \min_{v' \in B^{\mathcal{M}^{\theta}}_{\Delta_{\theta} \Plus}(v_{\theta})} w^T \bigcdot \mathbf{CF}(v', a;  \mathcal{M}^{\theta}) \geq b \Leftrightarrow 
	= \min_{\|\eta\|\leq \Delta_{\theta}}  w^T \bigcdot \mathbf{CF}(v, a;\mathcal{M}^{\theta}) + w^T \bigcdot S^{a}_{\theta}\eta \geq b \Leftrightarrow\\
	&= w^T \bigcdot \mathbf{CF}(v, a;\mathcal{M}^{\theta}) + \min_{\|\eta\|\leq \Delta_{\theta}}   w^T \bigcdot S^{a}_{\theta}\eta \geq b \Leftrightarrow
	= w^T \bigcdot \mathbf{CF}(v, a;\mathcal{M}^{\theta}) - \|w^T \bigcdot S^{a}_{\theta}\|^*\eta \geq b
	\end{aligned}
    \end{equation}
	The last equation is equivalent to the standard recourse problem for the classifier $h_{\theta}(x) = \text{sign}(w^T \bigcdot v - (b + \Delta \| w^T \bigcdot S^a_{\theta} \|^*))$. 
\end{proof}
\subsection{Proposition~\ref{pr:faroEx}}
\begin{proof}
By assumption the classifier $h(v)= \text{sign}(w^T \bigcdot v- b)$ and SCM $\mathcal{M}$ are linear. By Prop.~\ref{pr:optdec} a recourse action $a$ is adversarially fair robust if  for all $\theta \in \Theta_{\Delta}$ satisfies:
\begin{equation*}
	w^T \bigcdot v \geq b' \quad \forall v \in B^{\mathcal{M}^{\theta}}_{\Delta_{\theta} \Plus}(v_{\theta})
\end{equation*}
where $b' = b + \Delta \| w^T \bigcdot S^a_{\theta} \|^*$.
We employ the proof-by-contradiction method. Thus, we assume that no action exists that can shift the counterfactual perturbation to a favorable region.
By assumption, there exists a continuous feature $\mathbf{X}_j$ such that $\mathbf{X}_j$ is actionable and unbounded. Consider the middle action $a(v)$ that arbitrarily moves $x_j$. 
It does not matter whether this action is chosen as a hard or an additive intervention since the proof is the same in both cases. Thus we consider $a(v)=do(\mathbf{X}_j=\lambda)$ for $\lambda \in \mathbb{R}$ as a hard action. 
By Prop. \ref{pr:cfc}, $\mathbf{CF}(v, a;\mathcal{M}^{\theta}) = v + (\lambda -S_{\theta}^{-1}(v)_i) . (S_{\theta})_{*,i}$. 
By contradiction assumption, there exist $v \in B^{\mathcal{M}^{\theta}}_{\Delta_{\theta} \Plus}(v_{\theta})$ such that is not in favorable region so:
\begin{equation} \label{eq:lambda}
	\begin{aligned}
	& w^T \bigcdot \mathbf{CF}(v, a;\mathcal{M}^{\theta}) \leq b' \Leftrightarrow 
	w^T \bigcdot (v + (\lambda -S_{\theta}^{-1}(v)_i) . (S_{\theta})_{*,i}) \leq b' \Leftrightarrow \\
	& \lambda.w^T \bigcdot (S_{\theta})_{*,i}) \leq b'- w^T \bigcdot v + S_{\theta}^{-1}(v)_i.w^T
	\end{aligned}
\end{equation}
If $\lambda.w^T \bigcdot (S_{\theta})_{*,i})\neq0$, then clearly it is possible to set $\lambda$ to have arbitrarily large such that the inequality \ref{eq:lambda} is not valid. 
This event happens simultaneously for all $\theta$ because the mapping $S_{\theta}$ has a similar structure for all $\theta$.
Since $\mathbf{X}_j$ is actionable and unbounded, $a(x)=do(\mathbf{X}_j=\lambda)$ is a feasible action. Therefore, by contradiction, $a$ is a fair robust recourse action.
\end{proof}
\subsection{Corollary~\ref{pr:indifair}}
\begin{proof}
	Since $v$ does not belong to an unfair area, it is individually fair w.r.t the classifier, So it is reasonable to define the AFRR problem.
	By assumption, the feature $A$ is protected, so by means of reduction Lem. \ref{pr:decompose} for $\Delta \leq \Delta_0$, we can write:
	\begin{equation}
	r^{\mathcal{M}}_{\Delta}(v) = \max_{v' \in B^{\mathbf{CF}}_{\Delta}(v)} dist(v',L) = 
	\max_{a \in \mathcal{A}}\{\max_{v' \in B^{\mathcal{M}^{do(A = a)}}_{\Delta \Plus}(\ddot{v}_a)} dist(v',L)\} =
	\max_{a \in \mathcal{A}} \big\{ r^{\mathcal{M}^{do(\mathbf{A} = a)}}_{\Delta \Plus}(\ddot{v}_a) \big\} 
	\end{equation}
	In other hand by the Prop. \ref{pr:recCost} we have:
	$$r^{\mathcal{M}^{do(A = a)}}_{\Delta \Plus}(\ddot{v}_a) = \dfrac{|w^T\bigcdot \ddot{v}_a-b| + \Delta\|w^T.\ddot{\mathbf{S}}_a\|_{p^*}}{\|w\|_{p^*}}$$
	By reduction lemma, we can suppose $\mathcal{M}$ has one categorical variable that does not have parents. So without loss of generality, we suppose the index venerable $A$ in $V_i$ is $1$, so in this condition $\ddot{\mathbf{S}}_a = \mathbf{S}$. By substituting this fact in the last equation, the corollary is proved. 
\end{proof}
\subsection{Proposition~\ref{pr:delta0}}
\begin{proof}
	\begin{enumerate}[label=(\alph*)]
		\item
		By Corollary \ref{pr:indifair} we can write:
		\begin{equation*}
		\begin{aligned}
		\lim_{\Delta \rightarrow 0} r^{\mathcal{M}}_{\Delta}(v)&  = 
		\lim_{\Delta \rightarrow 0} \max_{a \in \mathcal{A}} \Big\{ \dfrac{|w^T\bigcdot \ddot{v}_a-b| + \Delta\|w^T.\mathbf{S}\|_{p^*}}{\|w\|_{p^*}}   \Big\} = 
		\max_{a \in \mathcal{A}} \bigg\{\lim_{\Delta \rightarrow 0}  
		\dfrac{|w^T\bigcdot \ddot{v}_a-b| + \Delta\|w^T.\mathbf{S}\|_{p^*}}{\|w\|_{p^*}}
		\bigg\} = \\ 
		& \max_{a \in \mathcal{A}} \{ \dfrac{|w^T\bigcdot \ddot{v}_a-b|}{\|w\|_{p^*}} \} =
		\max_{a \in \mathcal{A}} \Big\{  r^{\mathcal{M}}(\ddot{v}_a)  \Big\}
		\end{aligned}
		\end{equation*}
		\item
		If $\lim_{\Delta \rightarrow 0} r^{\mathcal{M}}_{\Delta}(v) = r^{\mathcal{M}}(v)$
		then $\displaystyle  \max_{a \in \mathcal{A}} \big\{  r^{\mathcal{M}}(\ddot{v}_a)  \big\} = r^{\mathcal{M}}(v)$ so we have individually fair recourse for instance $v$. Conversely if we have individual fairness at $v$ then $\displaystyle \max_{a \in \mathcal{A}} \big\{  r^{\mathcal{M}}(\ddot{v}_a)  \big\} = r^{\mathcal{M}}(v)$ it results $\lim_{\Delta \rightarrow 0} r^{\mathcal{M}}_{\Delta}(v) = r^{\mathcal{M}}(v)$.
	\end{enumerate}
\end{proof}
\subsection{Proposition~\ref{pr:faroProp}}
\begin{proof}
	If the limit $\displaystyle lim_{n \rightarrow \infty} a^*_{\Delta_n}$ exists, so for $\Delta_n$ the adversarially fair robust recourse has a solution. This fact support robustness of $v$. On the other hand part (a) of Prop. \ref{pr:delta0} $\displaystyle r_{\text{FARO}}^{\mathcal{M}} (v) = \lim_{\Delta \rightarrow 0} r_{\Delta}^{\mathcal{M}}(v) = \max_{a \in \mathcal{A}} \{r^{\mathcal{M}}(\ddot{v}_a)\}$ is invariant respect to twins, therefore, it is individual fair w.r.t recourse cost.
\end{proof}
\subsection{Proposition~\ref{pr:faroExist}}
\begin{proof}
	If for some $\Delta$, the solution AFRR problem exists then for each element of the monotone increasing sequence $(\Delta_n)^\infty_{ n =1}$ with condition $\lim \Delta_n \rightarrow 0$ such that $\Delta_n < \Delta$, the value of $r^{\mathcal{M}}_{\Delta_n}(v)$ exists, because the counterfactual perturbation $B^{\mathbf{CF}}_{\Delta_n} \subset B^{\mathbf{CF}}_{\Delta}$ so the AFRR problem has a solution by $B^{\mathbf{CF}}_{\Delta_n}$ with the cost value less than $r^{\mathcal{M}}_{\Delta}(v)$. Since the function $r^{\mathcal{M}}_{\Delta_n}(v)$ is monotone decreasing respect to $\Delta_n$, then by
	Monotone Convergence theorem the $\displaystyle \lim_{\Delta_n \rightarrow 0}  r^{\mathcal{M}}_{\Delta_n}(v)$ exists, therefore
	 for instance $v$ the value $r_{\text{FARO}}^{\mathcal{M}} (v)$ can be defined and correspondingly find the optimal action. 
\end{proof}
\section{How to Solve Challenges}
\label{app:solve_challenges}
 The ideas used to solve challenges are shown in the below table.
\begin{table}[H]
	\centering
	\begin{tblr}{ | p{0.5cm} |  p{14cm} |}
		\hline
		I & Define \hyperref[def:fairrec]{FARO recourse problem}. \\
		II &   
			$\circ$ Define \hyperref[def:cp]{counterfactual perturbation} based on metric instead of norm. \newline
			$\circ$ Consider \hyperref[rm:zero]{perturbation ball} $B^{d_{\mathcal{I},\mathcal{J}}}_{\Delta}(v)$ instead of ball around $0$.
			\\
		III & Define an \hyperref[def:midint]{middle intervention} to ensure that the counterfactual ball considers the categorical variable.\\
		V & Define \hyperref[def:fairrobust]{adversarially fair robust recourse problem}. \\
		\hline
	\end{tblr}
	\caption{The table of solutions provided for challenges.}
	\label{table:solutions}
\end{table}
\section{Simulation Details}
\label{app:simulation}
The structural equations used to generate the SCMs in \cref{sec:experiments} are listed below.
For the LIN and ANM SCMs, we generate the protected feature $A$ and variables $X_i$ according to the following structural equations:
\begin{itemize}
    \item linear SCM (LIN): 
\begin{equation}
\label{eq:lin_model}
\begin{cases}
A := U_A, & 			U_A \sim \mathcal{R}(0.5) 	\\
X_1 := 2A + U_1, & 	U_1\sim \mathcal{N}(0,1)	\\
X_2 := A-X_1 + U_2, & 			U_2 \sim \mathcal{N}(0,1)
\end{cases}
\end{equation}
    \item Additive Noise Model (ANM)
\begin{equation*}
\begin{cases}
A := U_A, & 			U_A \sim \mathcal{R}(0.5) 	\\
X_1 := 2A^2 + U_1, & 	U_1\sim \mathcal{N}(0,1)	\\
X_2 := AX_1 + U_2, & 			U_2 \sim \mathcal{N}(0,1)
\end{cases}
\end{equation*}
\end{itemize}
where $\mathcal{R}(p)$ is Rademacher random variables with probability $p$ and 
$\mathcal{N}(\mu,\sigma^2)$ is normal r.v. with mean $\mu$ and variance $\sigma^2$.
To generate Ground Truth, we use linear or non-linear methods for both aware and unaware baselines:
\begin{equation*}
h(A, X_1,X_2,X_3) = 
\begin{cases}
\textrm{sign}(A + X_1 + X_2 < 0)    &  \text{Linear and Aware}      \\ 
\textrm{sign}(X_1 + X_2 < 0)        &  \text{Linear and Unaware}    \\ 
\textrm{sign}((A + X_1 + X_2)^2 <2) &  \text{Non-Linear and Aware}\\
\textrm{sign}((X_1 + X_2)^2 <2)     & \text{Non-Linear and Unaware}
\end{cases}
\end{equation*}
We use the \citet{h2o.ai} package to train models and the \textit{h2o.grid} for tuning hyperparameters.
For GLM, we use \textit{alpha = seq(0, 1, 0.1)} with \textit{lambda\_search = TRUE}.
For SVM, we set \textit{gamma = 0.01}, \textit{rank\_ratio = 0.1}, and use a Gaussian kernel.
For GBM, we search for the optimal model among the following parameters: \textit{learn\_rate = c(0.01, 0.1)}, \textit{max\_depth = c(3, 5, 9)}, and \textit{sample\_rate = c(0.8, 1.0)}.
\section{Case Study Details}
\label{app:case_study}

The adult demographic dataset, which consists of more than 195,000 samples is a newer version of the Adult income data~\cite{ding2021retiring}. This data contains seven categorical attributes: 
\textbf{Class of worker} (COW: 9 levels), 
\textbf{marital status} (MAR:5 levels), 
\textbf{occupation} (Occupation: 529 levels),
\textbf{Place of birth} (POBP: 219 levels),
\textbf{Relationship to householder}: (RELP: 17 levels),
\textbf{Race code} (RAC1P: 9 levels) and
\textbf{SEX} (1 = Male, 2 = Female) which sex is protected variable.
Adult income also includes three continuous variable 
\textbf{age} (AGEP as an integer from 0 to 99), 
\textbf{Educational attainment} (SCHL as integer value from 1 to 24) and 
\textbf{hours per week} (WKHP: an integer from 1 to 99) where education and hours per week variables are actionable.

Furthermore, we consider a semi-synthetic SCM proposed by \citet{karimi2020algorithmic} that is based on a loan approval scenario. The data aims to reflect the intuitive relationships between variables in a practical loan approval process.
This semi-synthetic data consists of gender, age, education, loan amount, duration, 
income, and saving variables with the following structural equations and exogenous distributions:
\begin{align*}
\begin{cases}
G := U_G  & U_G \sim \text{Bernoulli}(0.5)\\
A := -35+U_A & U_A \sim \text{Gamma}(10, 3.5) \\
E := -0.5 + \bigg(1 + e^{-\big(-1 + 0.5 G + (1 + e^{- 0.1 A})^{-1} + U_E \big)}\bigg)^{-1}  & U_E \sim \N(0,0.25) \\
L := 1 + 0.01 (A - 5) (5 - A) + G + U_L & U_L \sim \N(0,4) \\
D := -1 + 0.1A + 2G + L + U_D & U_D \sim \N(0, 9)\\
I := -4 + 0.1(A + 35) + 2G + G E + U_I &  U_I \sim \N(0, 4)\\
S :=  -4 + 1.5 \mathbb{I}_{\{I > 0\}} I + U_S & U_S\sim\N(0, 25)
\end{cases}
\end{align*}
The labels $Y$ were generated using the following formula:
\begin{equation*}
Y\sim \text{Bernoulli}\left(\left(1+e^{-0.3(-L-D+I+S+IS)}\right)^{-1}\right).
\end{equation*}
\section{Additional Results}
\label{app:add_result}
\begin{figure}[ht]
\includegraphics[width=\textwidth]{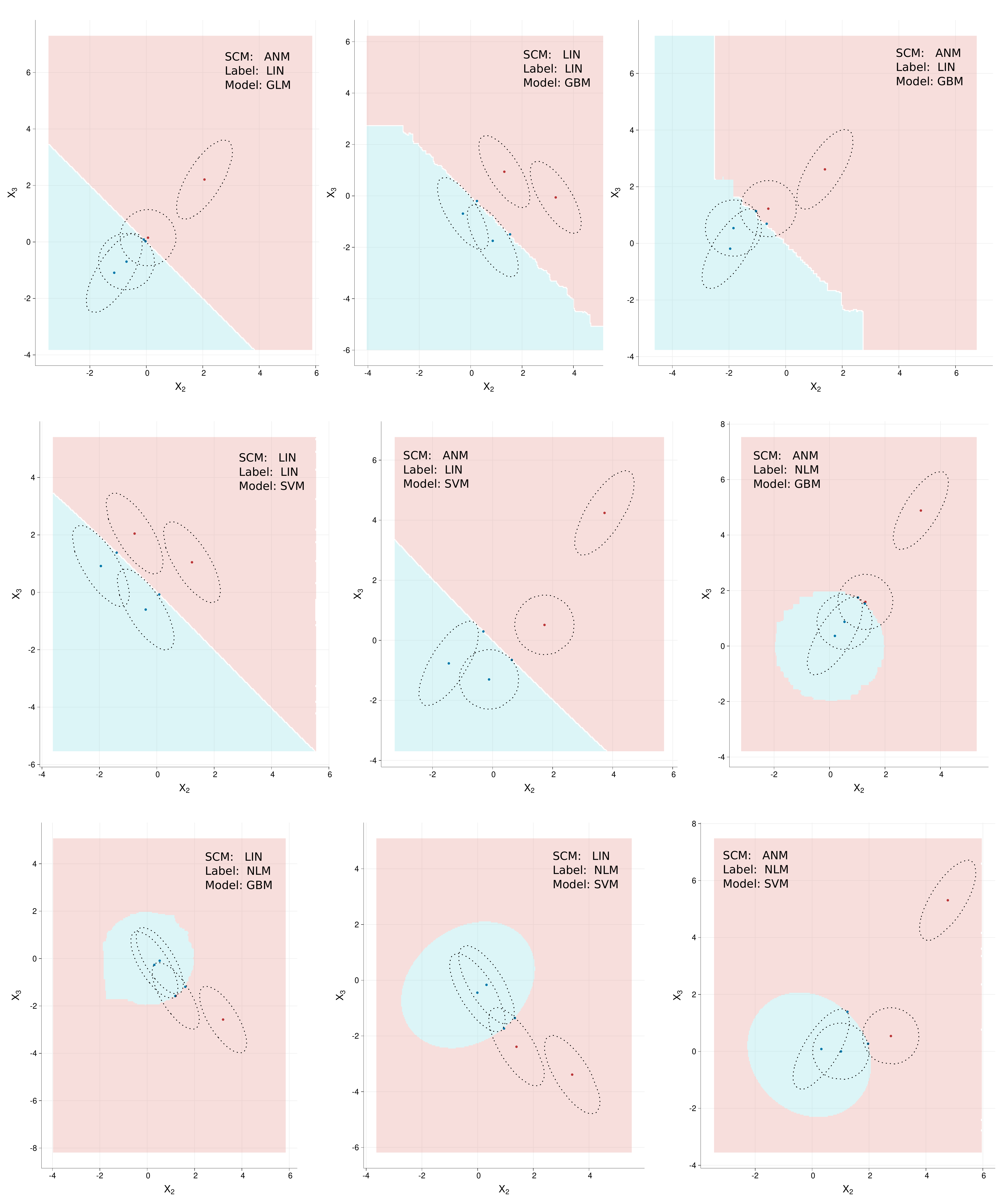}
\caption{Adversarial recourse in some simulations, with $\Delta=1$, for unaware labels and classifiers, including instances and their twins.}
\label{fig:counterfactual}
\end{figure}

In this appendix, we provide supplementary experimental results that were omitted from the main paper due to space limitations. We begin with a simulation of the results corresponding to the setup described in Section~\ref{sec:experiments}. The results for $\Delta = 1$ are shown in Table~\ref{tab:sim_1}, and the results for $\Delta = 0.5$ and $0.1$ will be presented in the following.
\begin{table}[ht]
\label{tab:sim_05}
\footnotesize
\begin{tabular}{ c l  c c c  |  c c c || c c c | c c c }
	\toprule
	\centering
	\multirow{3}{*}{\textbf{Classifier}}
	& \multicolumn{6}{r}{\textbf{GT labels from \textit{linear}}} 
	& \multicolumn{6}{r}{\textbf{GT labels from \textit{nonlinear}}}
	\\
	\cmidrule(r){3-8} \cmidrule(r){9-14}
	& \multicolumn{3}{r}{\textbf{LIN}} 
	& \multicolumn{3}{r}{\textbf{ANM}}
	& \multicolumn{3}{r}{\textbf{LIN}}
	& \multicolumn{3}{r}{\textbf{ANM}} 
	\\
	\cmidrule(r){3-5} \cmidrule(r){6-8} \cmidrule(r){9-11} \cmidrule(r){12-14}
	&
	& $\sigma_\textbf{R}$
	& $\sigma_\textbf{AR}$
	& $\sigma_\textbf{FR}$
	& $\sigma_\textbf{R}$
	& $\sigma_\textbf{AR}$
	& $\sigma_\textbf{FR}$
	& $\sigma_\textbf{R}$
	& $\sigma_\textbf{AR}$
	& $\sigma_\textbf{FR}$
	& $\sigma_\textbf{R}$
	& $\sigma_\textbf{AR}$
	& $\sigma_\textbf{FR}$
	\\
	\midrule
	\multirow{6}{*}{\rotatebox[origin=c]{90}{Aware Label}} 
& GLM$(A,X)$ & 0.71 & 0.6 & \textbf{0.00} & 1.14 & 1.03 & \textbf{0.00} & 1.8 & 1.36 & \textbf{0.00} & 2.49 & 2.08 & \textbf{0.00} \\
& SVM$(A,X)$ & 0.84 & 0.72 & \textbf{0.00} & 1.15 & 1.04 & \textbf{0.00} & 1.02 & 0.79 & \textbf{0.00} & 2.01 & 1.68 & \textbf{0.00} \\
& GBM$(A,X)$ & 1.22 & 1.01 & \textbf{0.00} & 1.13 & 1.02 & \textbf{0.00} & 1.16 & 0.86 & \textbf{0.00} & 1.69 & 1.43 & \textbf{0.00} \\
& GLM$(X)$ & 0.62 & 0.51 & \textbf{0.00} & 1.25 & 1.11 & \textbf{0.00} & 1.18 & 0.93 & \textbf{0.00} & 1.77 & 1.5 & \textbf{0.00} \\
& SVM$(X)$ & 0.79 & 0.65 & \textbf{0.00} & 1.1 & 0.98 & \textbf{0.00} & 0.92 & 0.71 & \textbf{0.00} & 1.66 & 1.42 & \textbf{0.00} \\
& GBM$(X)$ & 0.81 & 0.69 & \textbf{0.00} & 1.16 & 1.02 & \textbf{0.00} & 1.06 & 0.8 & \textbf{0.00} & 1.55 & 1.32 & \textbf{0.00} \\
	\midrule
    \multirow{6}{*}{\rotatebox[origin=c]{90}{Unaware Label}} 
& GLM$(A,X)$ & 0.59 & 0.46 & \textbf{0.00} & 1.19 & 1.06 & \textbf{0.00} & 1.52 & 1.12 & \textbf{0.00} & 2.4 & 1.99 & \textbf{0.00} \\
& SVM$(A,X)$ & 0.59 & 0.46 & \textbf{0.00} & 1.24 & 1.1 & \textbf{0.00} & 0.91 & 0.7 & \textbf{0.00} & 2.54 & 2.03 & \textbf{0.00} \\
& GBM$(A,X)$ & 0.75 & 0.6 & \textbf{0.00} & 1.24 & 1.08 & \textbf{0.00} & 1.04 & 0.77 & \textbf{0.00} & 1.77 & 1.49 & \textbf{0.00} \\
& GLM$(X)$ & 0.57 & 0.45 & \textbf{0.00} & 1.22 & 1.08 & \textbf{0.00} & 1.29 & 0.96 & \textbf{0.00} & 2.48 & 2.04 & \textbf{0.00} \\
& SVM$(X)$ & 0.57 & 0.45 & \textbf{0.00} & 1.3 & 1.14 & \textbf{0.00} & 0.89 & 0.69 & \textbf{0.00} & 2.12 & 1.74 & \textbf{0.00} \\
& GBM$(X)$ & 0.62 & 0.48 & \textbf{0.00} & 1.11 & 0.97 & \textbf{0.00} & 1.01 & 0.76 & \textbf{0.00} & 2.02 & 1.65 & \textbf{0.00} \\
\\
	\hline
\end{tabular}
\caption{Result table of comparison between classifiers with $\sigma_\textbf{R}$, r$\sigma_\textbf{AR}$, $\sigma_\textbf{FR}$ for $\Delta = 0.5$.}
\end{table}
\begin{table}[ht]
\label{tab:sim_01}
\footnotesize
\begin{tabular}{ c l  c c c  |  c c c || c c c | c c c }
	\toprule
	\centering
	\multirow{3}{*}{\textbf{Classifier}}
	& \multicolumn{6}{r}{\textbf{GT labels from \textit{linear}}} 
	& \multicolumn{6}{r}{\textbf{GT labels from \textit{nonlinear}}}
	\\
	\cmidrule(r){3-8} \cmidrule(r){9-14}
	& \multicolumn{3}{r}{\textbf{LIN}} 
	& \multicolumn{3}{r}{\textbf{ANM}}
	& \multicolumn{3}{r}{\textbf{LIN}}
	& \multicolumn{3}{r}{\textbf{ANM}} 
	\\
	\cmidrule(r){3-5} \cmidrule(r){6-8} \cmidrule(r){9-11} \cmidrule(r){12-14}
	&
	& $\sigma_\textbf{R}$
	& $\sigma_\textbf{AR}$
	& $\sigma_\textbf{FR}$
	& $\sigma_\textbf{R}$
	& $\sigma_\textbf{AR}$
	& $\sigma_\textbf{FR}$
	& $\sigma_\textbf{R}$
	& $\sigma_\textbf{AR}$
	& $\sigma_\textbf{FR}$
	& $\sigma_\textbf{R}$
	& $\sigma_\textbf{AR}$
	& $\sigma_\textbf{FR}$
	\\
	\midrule
	\multirow{6}{*}{\rotatebox[origin=c]{90}{Aware Label}} 
& GLM$(A,X)$ & 0.72 & 0.7 & \textbf{0.00} & 1.19 & 1.16 & \textbf{0.00} & 1.58 & 1.49 & \textbf{0.00} & 1.83 & 1.78 & \textbf{0.00} \\
& SVM$(A,X)$ & 0.81 & 0.78 & \textbf{0.00} & 1.05 & 1.03 & \textbf{0.00} & 0.96 & 0.91 & \textbf{0.00} & 2.06 & 1.98 & \textbf{0.00} \\
& GBM$(A,X)$ & 1.61 & 1.55 & \textbf{0.00} & 1.19 & 1.16 & \textbf{0.00} & 1.14 & 1.07 & \textbf{0.00} & 1.85 & 1.79 & \textbf{0.00} \\
& GLM$(X)$ & 0.63 & 0.61 & \textbf{0.00} & 1.15 & 1.12 & \textbf{0.00} & 1.16 & 1.1 & \textbf{0.00} & 2.56 & 2.45 & \textbf{0.00} \\
& SVM$(X)$ & 0.78 & 0.74 & \textbf{0.00} & 1.14 & 1.11 & \textbf{0.00} & 0.9 & 0.85 & \textbf{0.00} & 1.79 & 1.72 & \textbf{0.00} \\
& GBM$(X)$ & 0.93 & 0.87 & \textbf{0.00} & 1.1 & 1.07 & \textbf{0.00} & 0.96 & 0.9 & \textbf{0.00} & 1.79 & 1.72 & \textbf{0.00} \\
	\midrule
    \multirow{6}{*}{\rotatebox[origin=c]{90}{Unaware Label}} 
& GLM$(A,X)$ & 0.55 & 0.52 & \textbf{0.00} & 1.08 & 1.06 & \textbf{0.00} & 1.47 & 1.38 & \textbf{0.00} & 2.51 & 2.41 & \textbf{0.00} \\
& SVM$(A,X)$ & 0.55 & 0.52 & \textbf{0.00} & 1.16 & 1.13 & \textbf{0.00} & 0.85 & 0.8 & \textbf{0.00} & 2.31 & 2.21 & \textbf{0.00} \\
& GBM$(A,X)$ & 0.68 & 0.64 & \textbf{0.00} & 1.13 & 1.1 & \textbf{0.00} & 0.97 & 0.91 & \textbf{0.00} & 1.76 & 1.69 & \textbf{0.00} \\
& GLM$(X)$ & 0.58 & 0.55 & \textbf{0.00} & 1.27 & 1.24 & \textbf{0.00} & 1.51 & 1.4 & \textbf{0.00} & 2.33 & 2.23 & \textbf{0.00} \\
& SVM$(X)$ & 0.59 & 0.56 & \textbf{0.00} & 1.05 & 1.02 & \textbf{0.00} & 0.91 & 0.86 & \textbf{0.00} & 1.74 & 1.67 & \textbf{0.00} \\
& GBM$(X)$ & 0.88 & 0.82 & \textbf{0.00} & 1.2 & 1.16 & \textbf{0.00} & 0.97 & 0.91 & \textbf{0.00} & 2.01 & 1.92 & \textbf{0.00} \\
\\
	\hline
\end{tabular}
\caption{Result table of comparison between classifiers with $\sigma_\textbf{R}$, r$\sigma_\textbf{AR}$, $\sigma_\textbf{FR}$ for $\Delta = 0.1$.}
\end{table}
In Fig. \ref{fig:counterfactual}, the decision boundary, instances, their twins, counterfactual perturbations, and optimal robust actions can be found for various learning models and SCMs.

\end{document}